%% file: Ga_Ye_Co_Restless.tex
\documentclass[10pt,final,twocolumn]{IEEEtran}

\usepackage{textcomp}
\DeclareSymbolFont{tildelow}{TS1}{cmr}{m}{n}
\DeclareMathSymbol{\tildelow}{0}{tildelow}{126}

\usepackage{bbm}
\usepackage{xcolor}
\usepackage{color}
\usepackage{graphicx}
\usepackage{epsfig}
\usepackage{subfigure}
\usepackage{amssymb}
\usepackage{amsbsy}
\usepackage{cite}
\usepackage{caption}
\usepackage{amsthm}
\usepackage{romannum}
\usepackage[noend]{algorithmic}
\usepackage{algorithm}
\input{macros}

\usepackage[margin=0.95in]{geometry}

\usepackage{float}
\usepackage[T1]{fontenc}
\usepackage{url}
\usepackage{ifthen}
\usepackage{cite}
\usepackage[cmex10]{amsmath} 
\usepackage{tikz}

\renewcommand{\baselinestretch}{0.974}

\begin{document}

\title{Learning in Restless Bandits under Exogenous Global Markov Process}

\author{
  {Tomer Gafni, Michal Yemini, Kobi Cohen}
	\thanks{
		T. Gafni and K. Cohen are with the School of Electrical and Computer Engineering, Ben-Gurion University of the Negev, Beer-Sheva, Israel (e-mail:gafnito@post.bgu.ac.il; yakovsec@bgu.ac.il).} 
		\thanks{
	Michal Yemini is with the Department of Electrical and Computer Engineering, Princeton University, Princeton NJ, USA (e-mail: yemini.michal@gmail.com).
	}
	\thanks{A short version of this paper that introduces the algorithm, and preliminary simulation results was accepted for presentation in the IEEE International Conference on Acoustics, Speech and Signal Processing (ICASSP) 2022 \cite{gafni2022Exogenous}. 
	In this journal version we include: (i) A detailed
	discussion on the implementation of the algorithm; (ii) a deep theoretical analysis of the algorithm with detailed proofs; (iii) more extensive simulation results; and (iv) a detailed discussion of the results, and comprehensive discussion and comparison with the existing literature.}
	\thanks{This research was supported by the ISRAEL SCIENCE FOUNDATION
(grant No. 2640/20)}
	\vspace{-0.75cm}
}
\maketitle
\pagenumbering{arabic}

\begin{abstract}
\label{sec:abstract}
We consider an extension to the restless multi-armed bandit (RMAB) problem with unknown arm dynamics, where an unknown exogenous global Markov process governs the rewards distribution of each arm. Under each global state, the rewards process of each arm evolves according to an unknown Markovian rule, which is non-identical among different arms.
At each time, a player chooses an arm out of $N$ arms to play, and receives a random reward from a finite set of reward states. 
The arms are restless, that is, their local state evolves regardless of the player's actions. Motivated by recent studies on related RMAB settings, the regret is defined as the reward loss with respect to a player that knows the dynamics of the problem, and plays at each time $t$ the arm that maximizes the expected immediate value. The objective is to develop an arm-selection policy that minimizes the regret. To that end, we develop the Learning under Exogenous Markov Process (LEMP) algorithm. We analyze LEMP theoretically and establish a finite-sample bound on the regret. We show that LEMP achieves a logarithmic regret order with time. We further analyze LEMP numerically and present simulation results that support the theoretical findings and demonstrate that LEMP significantly outperforms alternative algorithms.
\end{abstract}


\section{Introduction}
\label{sec:introduction}
The multi-armed bandit (MAB) problem is a popular model for sequential decision making with unknown information: A player chooses actions repeatedly among $N$ different arms. After each action it receives a random reward having an unknown probability distribution that depends on the chosen arm. The objective is to maximize the expected total reward over a finite horizon of $T$ periods. Restless multi-armed bandit (RMAB) problems are generalizations of the MAB problem.
Differing from the classic MAB, where the states of passive arms remain frozen, in the RMAB setting, the state of each arm (active or passive) can change.
In this paper we consider an extension to the RMAB problem, in which we assume that an exogenous (global) Markov process governs the distribution of the restless arms, and thus the reward depends on both the state of the global process, and the local state of the chosen (active) arm.   

\subsection{Applications}

RMAB problems have attracted much attention for their wide application in diverse areas such as manufacturing systems \cite{sun2021dynamic}, economic systems \cite{scott2015multi}, biomedical engineering \cite{bulucu2019personalizing}, wireless communication systems \cite{hsu2019scheduling, gafni2022federated} and communication network \cite{xu2021task, amar2021online}. 
A particularly relevant application captured by the extended RMAB model considered in this paper is the Dynamic Spectrum Access (DSA) paradigm \cite{zhao2007survey, geirhofer2006measurement, wang2011optimality}, where primary users (licensed) occupy the spectrum occasionally, and a secondary user is allowed to transmit over a single channel when the channel is free. 
This behavior is commonly modeled by a Gilbert-Elliot model that comprises a Markov chain with two binary states. This model is captured as our exogenous (global) process, where global state $1$ denotes a transmitting primary user and global state $0$ denotes a vacant channel, i.e., an inactive primary user.
The statistical model for the arms establishes the relationship between a physical channel and its finite-state Markov model for a packet transmission system. We adopt the view of previous studies, forming a finite-state Markov channel model to reflect the fading channel effect \cite{wang1995finite}. 
The received SNR values are partitioned into a finite number of states according to a criterion based on the average duration of each state. This model is very useful and enables one to avoid slow bit-level simulations and focus on the overall system design. The differences in time scales of fading and primary user switching can be manifested in our model by increasing the probability of the global state to remain in its state, compared to the probability of a local state to remain in a certain state. 

Other potential applications of the model include important tasks in federated learning \cite{gafni2022federated}, recommendation systems \cite{elena2021survey, kulkarni2020context}, and dynamic pricing \cite{misra2019dynamic}.
Federated learning \cite{gafni2022federated} is an emerging machine learning paradigm for training models across multiple edge devices holding local datasets, without explicitly exchanging the data.
An important task in federated learning is to design user scheduling algorithms that determine which subset of users transmit at each round to save the communication resources, where the overall completion time of a task depends on the local processing time (including sensing, computation, transmission) of each selected user \cite{gafni2022federated}.
This application can be modeled by our new RMAB framework by modeling the random processing time of local users as Markov processes (e.g., due to Markovian channel fading, queue length of data samples, etc.). The global feature of the objective function in the federating learning task can be modeled by the global Markov process (e.g., the source state that generates the data samples, the location of the server that affects the communication channels, etc.). 

In recommendation system tasks \cite{elena2021survey, kulkarni2020context}, a well-known problem is the task of learning the "preference" or “rating” that a new user in the system would give to an item. The agent's goal is then to maximize the aggregated reward, which is associated with e.g., clicks, likes, shares, sells, etc. This problem is called the cold-start problem, and MAB-based formulations have been used to model it \cite{elena2021survey, kulkarni2020context}. Here, the arms represent items (or categories of items) and the reward distribution of each arm represents the user preference regarding this item. 
If the user interacts with the item, a positive feedback (i.e., a high reward) is given, if not, a negative feedback (e.g., a zero reward) is given. 
Furthermore, it is well known that often there are some global "trends" that affect the preferences of the user. Rather than assuming deterministic feature signals in MAB-based formulations \cite{kulkarni2020context}, using the new RMAB framework considered in this paper, global trends can be modeled by an exogenous global Markov process (drawn from an unknown Markovian process and needs to be learned). Our new model allows different distributions when arms are active (selected) or passive (not selected), which is well suited to capture different preference distributions when items are presented to the user or not. 

Finally, in dynamic pricing tasks \cite{misra2019dynamic}, a key problem for businesses is to learn online the market demands to set prices for products or services dynamically. Consider the pricing decision for a manager at an online retailer. At the time of pricing, the manager is unlikely to have complete information about each product’s demand curve. In these markets, a manager must consider an automated pricing policy to set real-time retail prices with incomplete demand information. For example, a consumer would buy a product only if its preference is higher than the price. Using our new RMAB model, the local arms represent the consumer's preferences which evolve as a Markov process. However, global economic changes (e.g., inflation) affect the consumer's preferences regarding various purchases. These global changes are captured by the global Markov process.  

\subsection{Performance Measure of RMAB under Exogenous Global Markov Process}
\label{ssec:performance_measure}

Computing the optimal policy for RMABs is P-SPACE hard even when the Markovian model is known \cite{papadimitriou1994complexity}, therefore, alternative tractable policies and objective functions have been proposed.
Nevertheless, always playing the arm with the highest expected reward is  optimal in the classic MAB under i.i.d. or rested Markovian rewards, up to an additional constant term \cite{anantharam1987asymptotically}.
Thus, a commonly used approach in classic RMAB (i.e., without exogenous process) with unknown dynamics settings to measure the algorithm performance in a tractable way defines the regret as the reward loss of the algorithm with respect to a genie that always plays the arm with the highest expected reward, also known as \textit{weak} regret \cite{tekin2012online,liu2013learning,gafni2020learning}.

However, in our setting, due to the exogenous process, each global state is associated with different "best" arm (i.e., the arm with the highest expected reward).
To accommodate the effect of the exogenous global Markovian state, we extend the definition of regret, and measure the performance of the algorithm by the reward loss of the algorithm with respect to a genie that plays in each time step the arm with the highest expected reward given the global state.
Furthermore, since the next global state is unknown before choosing the arm for the next time step, we adopt a myopic performance measure, as considered also in \cite{dai2011non,bagheri2015restless}.
That is, the objective in this paper is to select the arm that has the highest immediate \textit{expected} value at each time slot under unknown arm dynamics. 
The expected value, and thus also the arm selection, depend on both the transition probabilities of the global exogenous Markov process and the mean reward of the arms, which depends on the global state. 
Consequently, we define the regret as the reward loss of an algorithm with respect to a genie that knows the transition probabilities of the global process and the expected rewards of the local arms.
Thus, we note that the regret is not defined with respect to the best arm on average (that would result in a weak regret), but with respect to a strategy tracking the best arm at each step, which is stronger. 
This notion of regret was also considered in Section 8 of \cite{auer2002nonstochastic} and in \cite{garivier2011upper}, for the non-stochastic bandit problem.

\subsection{Main Results}
\label{ssec:results}

Due to the restless nature of both active and passive arms, learning the Markovian reward statistics requires
that arms will be played in a consecutive manner for a period of time (i.e., phase) \cite{liu2013learning, tekin2012online, gafni2020learning}.
Thus we divide the time horizon into two phases, an exploration phase and exploitation phase. The goal of the exploration phase is to identify the best arm for each global state before entering the exploitation phase. 

Upper Confidence Bound (UCB)-based policies, that are used to identifying the best arm, require parameter tuning depending on the unobserved hardness of the task \cite{audibert2010best,shahrampour2017sequential,shen2019universal}. 
The hardness parameter is a characteristic of the hardness of the problem, in the sense that it determines the order of magnitude of the sample complexity required to find the best arm with a required probability.
In the classic MAB formulation, the hardness of the task is characterized by $H_i = \frac{1}{(\mu^*-\mu^i)^2}$, where $\mu^*,\mu^i$ are the means of the best arm and arm $i$, respectively. 
However, since the hardness parameter is unknown, existing algorithms use an upper bound on $\max_i H_i$ (e.g., \cite{liu2013learning}), which increases the order of magnitude of exploration phases, and consequently the regret. Considering the above, we summarize our main results and contributions.

1) \emph{An Extended Model for RMAB:} RMAB problems have  been  investigated  under various models of observation distributions in past and recent years. The extended model considered in this paper is capable of capturing more complex scenarios and requires an adaptation of the regret measure as discussed above. Handling this extension in the RMAB setting leads to different algorithm design and analysis as compared to existing methods.  

2) \emph{Algorithm Development:} We develop a novel algorithm, dubbed Learning under Exogenous Markov Process (LEMP), that estimates online the appropriate hardness parameter from past observations (Sec.~\ref{ssec:motivation}). Based on these past observations, the LEMP algorithm generates adaptive sizes of exploration phases, designed to explore each arm in each global state with the appropriate number of samples. Thus, LEMP avoids oversampling bad arms, and at the same time identifies the best arms with sufficient high probability.
To ensure the consistency of the restless arms' mean estimation, LEMP performs regenerative sampling cycles (Sec.~\ref{ssec:exploration}). In the exploitation phases, LEMP dynamically chooses the best estimated arm, based on the evaluation of the global state (Sec.~\ref{ssec:exploitation}). 
The rules that decide when to enter each phase are adaptive in the sense that they are updated dynamically and controlled by the current sample means and the estimated global transition probabilities in a closed-loop manner (Sec.~\ref{ssec:selection}). 
Interestingly, the size of the exploitation phases is deterministic and the size of the exploration phases is random.

3) \emph{Performance Analysis:} We provide a rigorous theoretical analysis of LEMP algorithm. Specifically, we establish a finite sample upper bound on the expected regret, and show that its order is logarithmic with time. We also characterize the appropriate hardness parameter for our model (the $\overline{D}_i$ parameter defined in (\ref{eq:Local_Exploration_Rate})), and we demonstrate that estimating the hardness parameter indeed results in a scaled regret proportional to the hardness of the problem. The result in Theorem \ref{th:regret} also clarifies the impact of different system parameters (rewards, mean hitting times of the states, eigenvalues of the transition probability matrices, etc.) on the regret. We provide numerical simulations that support the theoretical results presented in this paper.

\subsection{Related Work}
\label{ssec:related}
The extended RMAB model considered here is a generalization of the classic MAB problem \cite{gittins1979bandit, lai1985asymptotically, anantharam1987asymptotically,auer2002finite, tekin2010online}. 
RMAB problems have been extensively studied under both the non-Bayesian \cite{tekin2012online, liu2013learning, gafni2020learning}, \cite{dai2011non,tekin2012approximately, xu2021online, karthik2021learning,gafni2019distributed,gafni2021distributed},  and Bayesian \cite{whittle1988restless, weber1990index, ehsan2004optimality,cohen2014restless}, \cite{zhao2008myopic,ahmad2009optimality,ahmad2009multi,liu2010indexability,wang2011optimality,wang2013optimality} settings.
Under the non-Bayesian setting, special cases of Markovian dynamics have been studied in \cite{tekin2012online}, \cite{dai2011non}. 
There are a number of studies that focused on special classes of RMABs. In particular, the optimality of the myopic policy was shown under positively correlated two-state Markovian arms \cite{zhao2007structure, zhao2008myopic,ahmad2009optimality,ahmad2009multi} under the model where a player receives a unit reward for each arm that was observed in a good state. In \cite{liu2010indexability}, \cite{liu2011indexability}, the indexability of a special classes of RMAB has been established.
In \cite{xu2021online}, the traditional restless bandit is extended by relaxing the restriction of a risk-neutral target function, and a general risk measure is introduced to construct a performance criterion for each arm.
Our work is also related to models of partially observed Markov decision process (POMDP) \cite{krishnamurthy2009partially}, with the goal of balancing between increasing the immediate reward and the benefits of improving the learning accuracy of the unknown states. 
In \cite{xu2021task}, the offloading policy design in a large-scale asynchronous MEC system with
random task arrivals, distinct workloads, and diverse deadlines is formulated as an RMAB problem, and the authors in \cite{krishnamurthy2001hidden} considered a tracking problem with independent objects and used an approximated Gittins index approach for finding policies. 

The setting in this paper is also related to the non-stationary bandit problems, where distributions of rewards may change in time \cite{garivier2011upper,hartland2006multi,yu2009piecewise,slivkins2008adapting,wang2018regional}.
However, the distribution that governs the non-stationary models in these studies differs from our settings, and leads to a different problem structure.
Finally, \cite{baltaoglu2016online, baltaoglu2016onlinea} and recently \cite{yemini2021restless} considered the setting of global Markov process that governs the reward distribution. However, they addressed the linear/affine model, and not the RMAB formulation that is explored in this paper. 
This new setting leads to fundamentally different algorithm design and regret analysis, mainly due to the restless nature of both active and passive arms in our model, that requires that arms will be played in a consecutive manner for a period of time.

\section{System Model and Problem Formulation}
\label{sec:problem}

    \begin{table} [htbp!]
    \centering
    \caption{Notations}
    \label{table:notation}
    \begin{tabular}{ |p{1cm}||p{6cm}|}
    \hline
     Notation& Description \\
    \hline \hline
    $N$ &  number of arms \\
    \hline
     $\mathcal{S}$ & global process\\
    \hline
    $s_t$ &  global state at time $t$ \\
    \hline
    $\mathcal{X}_s^i$ &  local process of arm $i$ in state $s$ \\
    \hline    
     $x_{s_t}^i$ &  reward (local state) of arm $i$ at global state $s_t$  \\
    \hline
     $\pi_s$ &  stationary distribution of state $s \in \mathcal{S}$ \\
    \hline
     $\pi_s^i(x)$ &  stationary distribution of local state $x$ in arm $i$ at global state $s$   \\
    \hline
     $p_{\tilde{s}\Check{s}}$ &  transition probability between states $\tilde{s}$ and $\Check{s}$ in $\mathcal{S}$\\
    \hline
     $\mu_s^i$ &  stationary reward mean of arm $i$ in state $s$\\
    \hline
     $V_s^i$ & expected value of arm $i$ in state $s$ (Eq.~(\ref{eq:Value}))\\
    \hline
    $\hat{V}_s^i(t)$ & estimated expected value of arm $i$ in state $s$ at time $t$ (Eq.~(\ref{eq:estimate_value}))\\
    \hline    
    $\overline{D}_s^i$ & exploration rate of arm $i$ in state $s$ (Eq.~(\ref{eq:Local_Exploration_Rate}))\\
    \hline  
    $\hat{D}_s^i(t)$ & estimated exploration rate of arm $i$ in state $s$ at time $t$ (Eq.~(\ref{eq:rate_estimate}))\\
    \hline    
    $T_s^i(t)$ & the number of samples from arm $i$ in global state $s$ in sub-block SB2 up to time $t$\\
    \hline     
    $t_s^i(n)$ & time index of the $n$th play on arm $i$ in global state $s$ in sub-block SB2\\
    \hline  
    $N_s(t)$ & number of occurrences of the state $s$ until time $t$\\
    \hline    
    $N_{s\tilde{s}}(t)$ & number of transitions from $s$ to $\tilde{s}$ up to time $t$.\\
    \hline    
    $I_L, I_G$ & local and global (respectively) minimal rate function (Eq.~(\ref{eq:I_L}), (\ref{eq:I_G}))\\
    \hline
    $L$ & exploration coefficient (Eq.~(\ref{eq:L}))\\
    \hline         
    \end{tabular}
    \end{table}

\begin{figure*}[htbp!]
    \centering
    \includegraphics[width=0.85\textwidth]{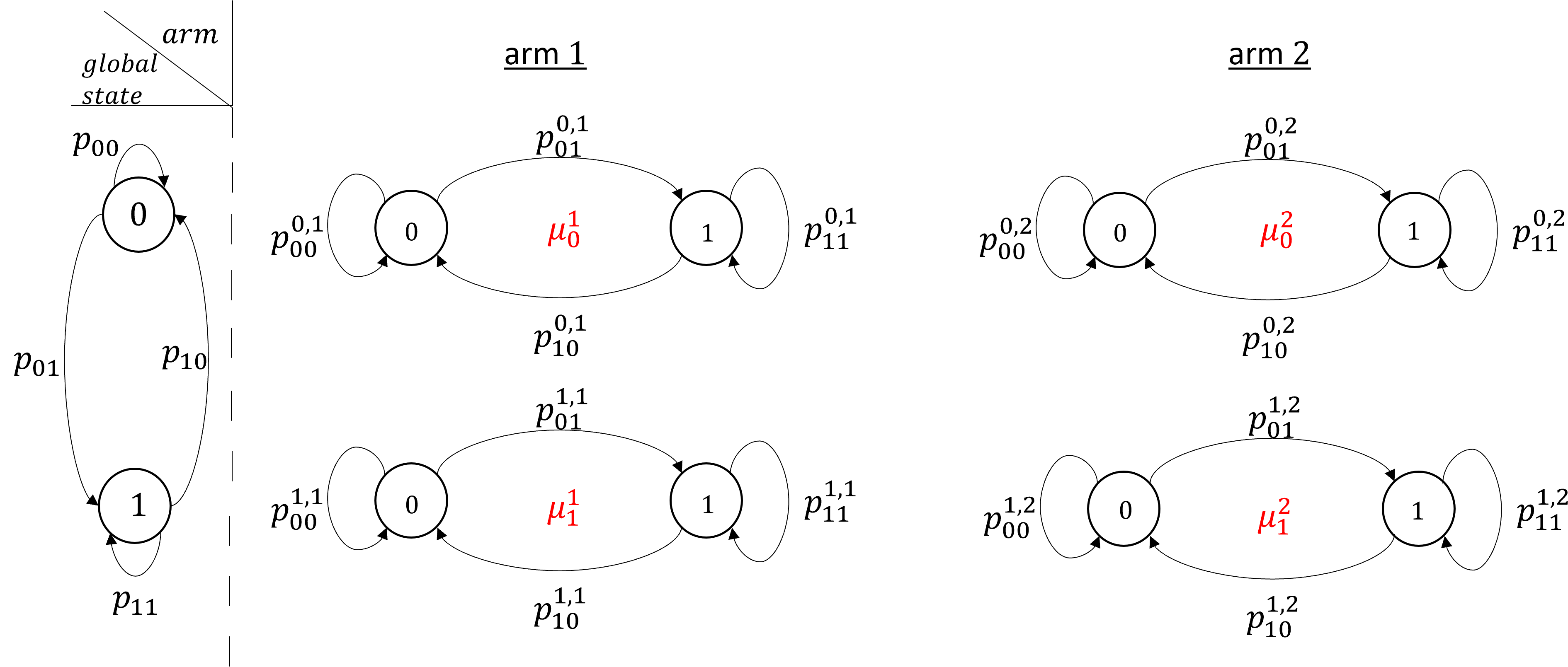}
    \vspace{-0.3cm}
    \caption{An illustration of the system model with $|\mathcal{S}|=2,N=2,|\mathcal{X}_s^i|=2, ~ \forall s,i$.}
    \vspace{-0.2cm}
    \label{fig:model_illustration}
\end{figure*}

We consider a set of $N$ arms, indexed by $\{1,\ldots,N\} \triangleq \mathcal{N}$, and a global system state process $\{s_t\}_{t=1,2,...}$, which is governed by a finite space, irreducible, and aperiodic discrete time Markov chain $\mathcal{S}$ with unknown transition matrix $P_S$.
We denote the transition probability between states $\tilde{s}$ and $\Check{s}$ in $\mathcal{S}$ by $p_{\tilde{s}\Check{s}}$, and we denote by $\pi_s$ the stationary distribution of states $s \in \mathcal{S}$.
For each global state $s \in \mathcal{S}$, the $i^{th}$ arm is modeled as a finite space, irreducible, and aperiodic discrete time Markov chain $\mathcal{X}^i_s$ with unknown transition matrix $P_{\mathcal{X}_s^i}$. 
We denote the transition probability between states $x$ and $y$ in $\mathcal{X}^i_s$ by $p_{xy}^{s,i}$.
We assume that $\mathcal{X}^i_{\tilde{s}} \bigcap \mathcal{X}^i_{\check{s}} = \emptyset$ for all $i, \tilde{s},\check{s}$ (i.e., we can recover the global state in each time slot \footnote{If this assumption does not hold, the player may recover the global state using techniques for the specific application task, e.g., spectrum sensing methods in the DSA paradigm \cite{al2018accurate}.}).
We also define the stationary distribution of state $x$ in arm $i$ at global state $s$ to be $\pi_s^i(x)$.
An illustration for the model with $|\mathcal{S}|=2,N=2,|\mathcal{X}_s^i|=2, ~ \forall s,i$ is given in Fig.~\ref{fig:model_illustration}.

At each time $t$, the player chooses one arm to play. When played, each arm offers a certain positive reward that defines the current state of the arm, $x^i_{s_t}$. 
The player receives the reward of the chosen arm, and infers the current global state $s_t$. Then, the global state transitions to a new state, which is unknown to the player before choosing the next arm to play.
We assume that the arms are mutually independent and restless, i.e., the local states of the arms continue to evolve regardless of the player’s actions according to the unknown Markovian rule $P_{\mathcal{X}_s^i}$.
The unknown stationary reward mean of arm $i$ at global state $s$, $\mu_s^i$, is given by:
\begin{center}
    $\mu_s^i = \sum \limits_{x \in \mathcal{X}^i_s} x \pi_s^i(x)$.
\end{center}
We further define the expected value of arm $i$ in global state $s$ to be
\begin{equation}
    \label{eq:Value}
    V_s^i \triangleq \sum \limits_{\check{s} \in \mathcal{S}} p_{s \check{s}} \mu_{\check{s}}^i.
\end{equation}
Let $\sigma$ be a permutation of $\{1,...,N\}$ such that
\begin{center}
$V_s^{\sigma(1)} \geq V_s^{\sigma(2)} \geq \cdots \geq V_s^{\sigma(N)}$.
\end{center}
Let $V_{s_t}^i(t)$ denote the value of arm $i$ at time $t$, let $i_t^*$ be the arm with the highest expected value at time $t$, i.e., $i_t^* \triangleq \arg \max_i V_{s_t}^i(t)$, and let $\phi(t)\in\left\{1, 2, ..., N\right\}$ be a selection rule indicating which arm is chosen to be played at time~$t$, which is a mapping from the observed history of the process to $\mathcal{N}$. Denote \[V_{\text{e}}(n)=\{i:V_{s_n}^i(n) < V_{s_n}^{\sigma(1)}(n)\}.\]
The expected regret of policy $\phi$ is defined as: 
\beq
\label{eq:regret_definition}
\hspace{-0.0cm}
\mathbb{E}_{\phi}[r(t)]= \mathbb{E}_{\phi}\bigg[\sum \limits_{n=1}^{t} \sum \limits_{i\in V_{\text{e}}(n)} (x^{i^*_n}_{s_n}(n) - x^i_{s_n}(n))\mathbbm{1}_{\{\phi(n)=i\}}\bigg],
\eeq
where hereafter $\mathbbm{1}_{\{A\}}$ denotes an indicator of an event $A$.
The objective is to find a policy that minimizes the growth rate of the regret with time (this notion of regret is similar to the “regret against arbitrary strategies” introduced in Section 8 of \cite{auer2002nonstochastic} and in \cite{garivier2011upper} for the non-stochastic bandit problem).
We note that, in this paper, the regret is not defined with respect to the best arm on average (e.g., as in RMAB models in \cite{tekin2012online,liu2013learning,gafni2020learning}), but with respect to the best arm at each step according to the instantaneous global state, which is a stronger regret.

\section{The Learning under Exogenous Markov Process (LEMP) Algorithm}
\label{sec:Algorithm 1}
The LEMP algorithm divides the time horizon into two types of phases, namely exploration and exploitation. In order to ensure sufficient small regret in exploitation phases (i.e., to reduce the probability for choosing sub-optimal arms in exploitation), our strategy estimates the required exploration rate of each arm, and updates the arm selection dynamically with time, controlled by the random sample means and transition probability estimates in a closed loop manner.    

\begin{algorithm}
\footnotesize
\label{alg:Algorithm}
\caption{LEMP Algorithm}
  \begin{algorithmic}
    \STATE   initialize: $\displaystyle t=0,N_s=0,n_I=0,n_O^i=1,T_s^i=0, r_s^i=0,  \forall               i=1  \ldots  N$ 
    \FOR{i=1:N}
      \STATE     play arm $i$; observe global state $s$ (denote the previous global state by $\tilde{s}$) and local state as $x$ and set $\gamma^i(n_O^i)=x$
      \STATE    $t:=t+1$; $T_s^i:=T_s^i+1$; $n_O^i:=n_O^i+1$; $r_s^i:=r_s^i+r_{x_s}$ $N_s=N_s+1;$ 
    \ENDFOR
    \WHILE{(1)}
     \FOR{$i=1:N$}
      \STATE    set $\hat{\mu}_s^i(t)$,$\hat{p}_{\tilde{s} s}(t)$,$\widehat{D}_s^i(t)$ according to (\ref{eq:mean_estimate}),(\ref{eq:transition_estimate}),(\ref{eq:rate_estimate}), respectively;  
     \ENDFOR    
     \WHILE{ condition (\ref{eq:condition1}) holds for some arm $i$ (or condition (\ref{eq:condition2}) holds)}
        \STATE   play arm $i$ (or arm $i_M$); observe global state $s$ and local state as $x$; 
        \WHILE{$x \neq \gamma^i(n_O^i-1)$ (SB1)} 
          \STATE            $t:=t+1$ $N_s=N_s+1$
          \STATE             play arm $i$; observe global state $s$ and local state as $x$
        \ENDWHILE
        \STATE         $t:=t+1;  T_s^i:=T_s^i+1;  r_s^i:=r_O^i+r_{x_s};N_s=N_s+1$; 
        \FOR{$n=1:4^{n_O^i-1}$ (SB2)}
          \STATE             play arm $i$; observe global state $s$ and local state as $x$
          \STATE            $t=t+1;  T_s^i=T_s^i+1;  r_s^i=r_s^i+r_{x_s}; N_s=N_s+1$;  
        \ENDFOR 
        \STATE    $n_O^i:=n_O^i+1$; set $\hat{\mu}_s^i(t)$,$\hat{p}_{\tilde{s} s}(t)$ according to (\ref{eq:mean_estimate}),(\ref{eq:transition_estimate}),(\ref{eq:rate_estimate}), respectively;  
        $\gamma^i(n_O^i)=x$
      \ENDWHILE
      \STATE set $\hat{V}_s^i(t) \; \forall s$ according to (\ref{eq:estimate_value})
      \STATE set   $i_s^* = \arg \max_i \hat{V}_s^i(t) \; \forall s$ 
      \FOR{$n=1:2 \cdot 4^{n_I-1}$}
        \STATE         play arm $i_s^*$; observe new global state $s$ and local state as $x$
        \STATE            $t:=t+1; N_s = N_s+1$
      \ENDFOR
      \STATE        $n_I:=n_I+1$; 
    \ENDWHILE
  \end{algorithmic}
\end{algorithm}

\subsection{Design principles of LEMP}
\label{ssec:motivation}
For sufficient small regret during exploitation phases, we should take a sufficiently large number of samples in the exploration phases. From (\ref{eq:Value}) we observe that we should estimate accurately two terms: the mean reward of each arm $i$ in each global state $s$, $\mu_s^i$, and the transition probabilities of the global Markov chain $\mathcal{S}$, $p_{\check{s} \tilde{s}}$. 

In the analysis, we show that in each global state $s$, we must explore a suboptimal arm $i$ with a \textit{local exploration rate} of at least $\overline{D}_s^i \log(t)$ times for being able to distinguishing it from $i_s^* \triangleq \arg \max_i V_s^i$ (i.e., the arm that maximizes the expected value in state $s$) with a sufficiently high accuracy, where 
\begin{equation}
\label{eq:Local_Exploration_Rate}
    \overline{D}_s^i \triangleq \frac{4L}{(V_s^* - V_s^i)^2},
\end{equation}
where $V_s^* \triangleq \max_i V_s^i$, and $L$ is the exploration coefficient that depends on the system parameters, defined in (\ref{eq:L}).
The $\overline{D}_s^i$ parameter is a type of hardness parameter \cite{audibert2010best}, appropriate for the setting considered in this paper, in the sense that it determines the order of magnitude of the sample size required to find the best arm in each global state with a required probability.

We point out that in order to derive $\overline{D}_s^i$, 
we should know the system parameters $\left\{p_{s\check{s}}\right\},\left\{\mu_{\check{s}}^i\right\}$.
Since the reward means and the transition probabilities are unknown, we estimate $\overline{D}_s^i$ by replacing $\mu_s^i,p_{s \check{s}}$ by their estimators: 
\begin{equation}
\label{eq:mean_estimate}
    \hat{\mu}_s^i(t) = \frac{1}{T_s^i(t)} \sum \limits_{n=1}^{T_s^i(t)} x^i_s(t_s^i(n)),
\end{equation}
\begin{equation}
\label{eq:transition_estimate}
    \hat{p}_{s \check{s}}(t) = 
\frac{N_{s \check{s}}(t)}{N_{s}(t)}.
\end{equation}
where $t_s^i(n)$ is the time index of the $n^{th}$ play on arm $i$ in global state $s$ in sub-block SB2 only (SB2 is detailed in Sec.~\ref{ssec:exploration}), $T_s^i(t)$ is the number of samples from arm $i$ in global state $s$ in sub-block SB2 up to time $t$, $N_s(t)$ is the number of occurrences of the state
$s$ until time $t$, and $N_{s \check{s}}(t)$ is the number of transitions from $s$ to $\check{s}$ up to time $t$.
We also define: $\Delta_s^i \triangleq (V_s^*-V_s^i)^2$, $\Delta_s \triangleq \min_i \Delta_s^i$, and we define $0<\Delta \leq \min_s \Delta_s$ to be a known lower bound on $\min_s \Delta_s$.

Denote the estimator of $\overline{D}_s^i$ by:
\begin{equation}
\label{eq:rate_estimate}
\hat{D}_s^i(t) \triangleq \frac{4L}{\max\{\Delta,(\hat{V}_s^*(t)-\hat{V}_s^i(t))^2 - \epsilon\}}, 
\end{equation}
where:
\begin{equation}
    \label{eq:estimate_value}
    \hat{V}_s^i(t) \triangleq \sum \limits_{\check{s} \in \mathcal{S}} \hat{p}_{s \check{s}}(t) \hat{\mu}_{\check{s}}^i(t),
\end{equation}
$\hat{V}_s^*(t) \triangleq \max_{i} \hat{V}_s^i(t)$, $\epsilon$ > 0 is a fixed tuning parameter, and $L$ is defined in (\ref{eq:L}).

Using $\{\hat{D}_s^i(t)\}$, which are updated dynamically over time and controlled by the corresponding estimators, we can design an adaptive arm selection for sampling arm $i$ at state $s$ that will converge to its exploration rate required for efficient learning, as time increases.
That is, the LEMP algorithm generates adaptive sizes of exploration phases, designed to explore each arm in each global state with the appropriate number of samples. Thus, LEMP avoids oversampling bad arms, and at the same time identifies the best arms with sufficiently high probability.
Whether we succeed to obtain a logarithmic regret order depends on how fast $\hat{D}_s^i(t)$ converges to a value that is no smaller than $\overline{D}_s^i$ (so that we take at least $\overline{D}_s^i$ samples from bad arms in most of the times).

\begin{figure*}[htbp]
    \centering
    \includegraphics[width=0.8\textwidth]{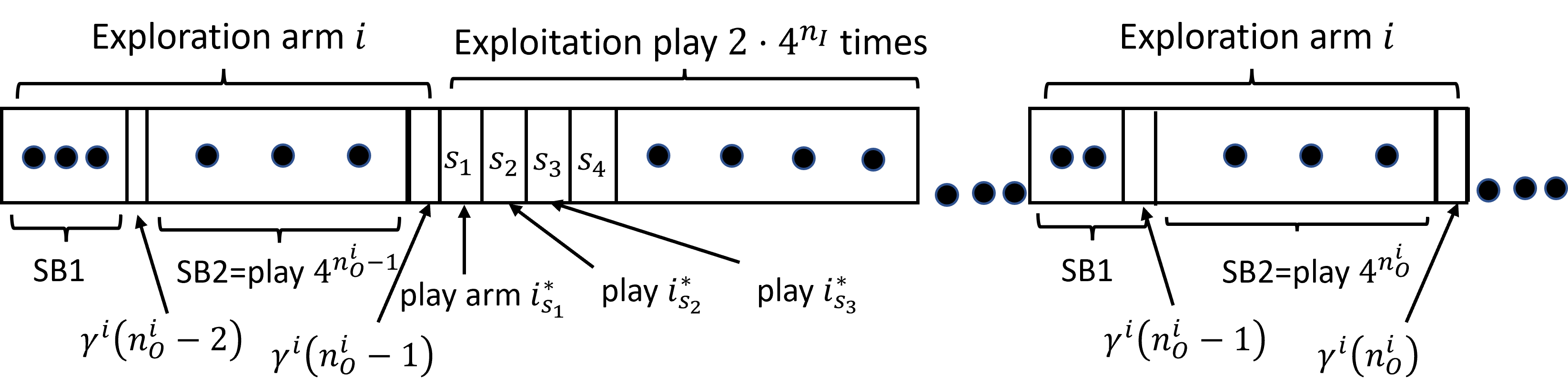}
    \vspace{-0.2cm}
    \caption{An illustration of the exploration and exploitation phases of LEMP Algorithm.}
    \vspace{-0.3cm}
    \label{fig:illustration}
\end{figure*}

\subsection{Description of the exploration phases:}
\label{ssec:exploration}
Due to the restless nature of both active and passive arms, learning the Markovian reward statistics requires that arms will be played in a consecutive manner for a period of time (i.e., phase).
Therefore, the exploration phases are divided into sub-blocks SB1 and SB2.
Consider time $t$ (and we remove the time index $t$ for convenience). 
We define $n_O^{i}(t)$ as the number of exploration phases in which arm $i$ was played up to time $t$.
Let $\gamma^{i}(n_O^{i}-1)$ be the last reward state observed at the $(n_O^{i}-1)^{th}$ exploration phase for arm $i$. As illustrated in Fig.~\ref{fig:illustration}, once the player starts the $(n_O^{i})^{th}$ exploration phase, it first plays a random period of time, also known as a random hitting time, until observing state $\gamma^i(n_O^{i}-1)$. This random period of time is referred to as SB1. Then, the player plays arm $i$ until it observes $4^{n_O^{i}}$ samples. 
This period of time is referred to as SB2. The player stores the $(4^{n_O^{i}})^{th}$ state $\gamma^i(n_O^{i})$ observed at the current $(n_O^{i})^{th}$ exploration phase, and so on. We define the set of time indices during SB2 sub-blocks by $\mathcal{V}_i$. This procedure ensures that each interval in $\mathcal{V}_i$ starts from the last state that was observed in the previous interval. Therefore, cascading these intervals forms a sample path which is equivalent to a sample path generated by continuously sampling the Markov chain. \vspace*{-0.1cm}

\subsection{Description of the exploitation phases:}
\label{ssec:exploitation}
Let $n_I(t)$ be the number of exploitation phases up to time $t$.
The player plays the exploitation phase for a deterministic period of time with length $2\cdot4^{n_I(t)-1}$ according to the following rule: at each time slot the player computes the expected value, $\hat{V}_s^i(t)$, of each arm given the observed global state when entering the $(n_I)^{th}$ exploitation phase, and plays the arm that maximizes the expected value.

\subsection{Phase Selection Conditions:}
\label{ssec:selection}

At the beginning of each phase, the player needs to decide whether to enter an exploration phase for one of the $N$ arms, or whether to enter an exploitation phase. We recall that the purpose of the exploration phases is to estimate both the expected rewards of the arms, and the transition probabilities of the global process.
We therefore define: 
\begin{equation}
\label{eq:I_L}
    \displaystyle I_L \triangleq \frac{ \Bar{\lambda}_{\min}}{3072 ((x_{\max}+2)^2 \cdot |\mathcal{X}_{\max}| \cdot \hat{\pi}_{\max} \cdot  |\mathcal{S}| \cdot (V_{\max}^* +2) )^2},
\end{equation}
\begin{equation}
\label{eq:I_G}
\displaystyle I_G \triangleq \frac{1 }{128 \big( (x_{\max}+2) \cdot |S| \cdot (V_{\max}^* +2) \big)^2 },
\end{equation}
which we denote as the local and global (respectively) minimal rate functions.
The decision to explore or exploit will be made due to the next two conditions:
first, if there exists an arm $i$ and a global state $s$ such that the following condition holds:
\beq
\label{eq:condition1}
T_s^i(t) \leq\max\left\{\hat{D}_s^i(t), \frac{2}{\epsilon^2 \cdot I_L}\right\}\cdot\log t,
\eeq
then the player enters an exploration phase for arm $i$.
Second, if there exists a global state $s \in \mathcal{S}$ where
\beq
\label{eq:condition2}
N_s(t) \leq \frac{2}{\epsilon^2 \cdot I_G} \cdot\log t,
\eeq
then the player enters an exploration phase for arm $i_M$ where $i_M \triangleq \arg \min_i \{ \min_s \hat{D}_s^i(t)\}$.
Otherwise, the player enters an exploitation phase. 
This selection rule of the LEMP algorithm is thus designed based on the following insights. First, the algorithm must take at least $\overline{D}^i_s(t) \cdot \log(t)$ samples from each sub-optimal arm for computing a sufficiently accurate estimate of the expected value $\hat{V}^i_s (t)$. Since $\overline{D}^i_s(t)$ depends on the expected values which are unknown, the algorithm replaces the unknown value $\overline{D}^i_s(t)$ by $\hat{D}^i_s(t)$, which overestimates $\overline{D}^i_s(t)$ to obtain the desired property. Second, since $\hat{D}^i_s(t)$ is a random variable, we need to make sure that the desired property holds with a sufficiently high probability. The parameters $I_L$, $I_G$ in (\ref{eq:condition1}), (\ref{eq:condition2}) are used to guarantee the desired property.

\section{Regret Analysis}
\label{sec:regret}
In the following theorem we establish a finite-sample bound on the expected regret as the function of time, resulting in a logarithmic regret order. 

\begin{theorem}
\label{th:regret}
Assume that LEMP algorithm is implemented and the assumptions on the system model described in Section \ref{sec:problem} hold, and an upper bound on $\Delta$ in known. Let $\lambda_s^i$ be the second largest eigenvalue of $P_{\mathcal{X}_s^i}$, and let $M^{s,i}_{x,y}$  be the mean hitting time of state $y$ starting at initial state $x$ for arm $i$ in global state $s$.
Define $x_{\max} \triangleq \max_{s\in \mathcal{S}, i\in \mathcal{N}} x_s^i$, $|\mathcal{X}_{\max}| \triangleq \max_{s\in \mathcal{S}, i\in \mathcal{N}} |\mathcal{X}_s^i|$ , $\pi_{\min}\triangleq \min_{s \in \mathcal{S},i\in \mathcal{N}, x \in \mathcal{X}_s^i} \pi^i_s(x)$, 
$\hat{\pi}_{\max} \triangleq \max_{s \in \mathcal{S}, i\in \mathcal{N}, x \in \mathcal{X}_s^i } \{\pi^i_s(x),1-\pi^i_s(x) \}$, $V_{\max}^* = \triangleq \max_{s \in \mathcal{S}} V_s^*$,
$\lambda_{\max}\triangleq\max_{s \in \mathcal{S},i\in \mathcal{N}}\ \lambda_s^i$,
$\overline{\lambda}_{\min}\triangleq 1-\lambda_{\max}$,
$\overline{\lambda_s^i}\triangleq 1-\lambda_s^i$,
$ M^i_{s,max}\triangleq\max_{x,y \in \mathcal{X}_s^i, x\neq y}M^{s,i}_{x,y}, M_{\max}^i \triangleq\max_s M^i_{s,max}$,
\begin{equation}
\label{eq:L}
   \hspace{-0.4cm} \displaystyle L \geq \frac{1}{16(V_{\max}^*+2)^2} \cdot \max \left\{\frac{1}{I_L},\frac{1}{I_G} \right\}.
\end{equation}
Then, the regret at time $t$ is upper bounded by:
\beq
\bea{l}
\label{eq:regret}
\displaystyle \mathbb{E}_{\phi} [r(t)] \leq x_{\max} \cdot \bigg[\sum \limits_{i=1}^N 
\big(\frac{1}{3} [4(3A_i\cdot \log(t)+1)-1]\\
\vspace{0.3cm} \hspace{3cm}  +M^i_{max} \cdot \log_4(3A_i\log(t)+1) \big) \\ \hspace{-0.2cm}
+6 N |\mathcal{S}| (\frac{|\mathcal{S}||\mathcal{X}_{\max}|}{\pi_{\min}}+2|\mathcal{S}|) \max_s \pi_s \cdot \lceil \log_4(\frac{3}{2}t+1) \rceil
\bigg] \\ \hspace{-0.2cm}
+\beta,
\ena
\eeq
where
\begin{equation}
\label{eq:A_i}
\hspace{-0.0cm} A_i\triangleq
\left\{ \begin{matrix}
\max\{\frac{2}{\epsilon^2 I_L}\;,\frac{2}{\epsilon^2 I_G}\;, \;\displaystyle\max_s\overline{D}_{s,max}^i\} \;, & \mbox{if $\forall s: i\in\mathcal{K}_s$}    \vspace{0.2cm}\\
\max \{\frac{2}{\epsilon^2 I_L}\;,\frac{2}{\epsilon^2 I_G}\;,\;4L/\Delta\} \;, & \mbox{if $ \exists s: i\nin\mathcal{K}_s$}
\end{matrix} \right.,
\end{equation}
$\displaystyle\overline{D}_{s,max}^i \triangleq{\frac{4L}{(V_s^* - V_s^i)^2-2\epsilon}}$, $\mathcal{K}_s$ is defined as the set of all indices $i\in{\{2,...,N\}}$ in global state $s$ that satisfy:
\begin{center}
$\displaystyle (V_s^* - V_s^{\sigma(i)})^2-2\epsilon > \Delta_s$, 
\end{center}
and 
\begin{center}
$\beta \leq 8 |\mathcal{S}| N \frac{|\mathcal{X}_{max}|}{\pi_{min}} \cdot
\left[\zeta(2+ \delta)+\frac{1}{1+\delta}\zeta(1+ \delta)\right]$,
\end{center}
for some arbitrarily small $\delta>0$, where $\zeta(\cdot)$ is the Reimann zeta function.
\end{theorem}

Theorem \ref{th:regret} shows that the regret under LEMP has a logarithmic order with time. The scaling of the regret with the mean hitting time $M^i_{max}$ under LEMP (which arises due to the construction of SB1 in the exploration phases) is of order $O(\sum_i M_{max}^i\log\log t)$. We point out that the scaling with $M^i_{max}$ under LEMP is significantly better than the scaling under other algorithms that use regenerative cycles. For example, in \cite{tekin2012online}, the RCA algorithm performs random regenerative cycles until catching predefined states in \emph{each phase}, thus the scaling with the mean hitting time (which scales at least polynomially with the state space) is $O(\sum_i M_{max}^i\log t)$.
The scaling with $N$ and $\Delta$ under LEMP is of order
$O\left((\frac{1}{\sqrt{\Delta}}+N-2)\log t\right)$ since every bad arm is sampled according to its unique exploration rate which is estimated by the adaptive sequencing rules. We point out that the scaling with $N$ and $\Delta$ under LEMP is significantly better than the scaling under other algorithms that divides the time horizon into exploration and exploitation phases. For example, in \cite{liu2013learning}, the scaling under the DSEE algorithm is of order $O\left((\frac{1}{\sqrt{\Delta}}+\frac{N-2}{\Delta})\log t\right)$ since all bad arms are explored according to the worst exploration rate. 
We also point out that the scaling of the regret under LEMP with the size of the global space $|\mathcal{S}|$ and local space $|\mathcal{X}_{\max}|$ is similar to the scaling of these parameters in classic works in the literature (e.g,. \cite{auer2008near} for $|\mathcal{S}|$ and \cite{tekin2010online,tekin2012online} for $|\mathcal{X}_s^i|$).
Moreover, if the exogenous Markov process has just a single state, we obtain the typical regret bound for restless Markovian MAB (e.g., as in \cite{gafni2020learning}), as expected. 
However, we note that if the local arms evolve as an i.i.d process, we do not obtain the bound of stochastic $K$-armed bandit. This is due to the fact that under the restless Markovian model we are required to design a much more complicated algorithm compared to algorithms in the stochastic $K$-armed bandit (e.g., UCB1) in order to obtain the logarithmic regret.

We finally note that the model considered in this paper can alternatively be adjusted such that we allow transitions of local arms at each time unit, and allow transitions of the global arm at each $K$ time units, as done, for example, in queuing system applications, when some servers may be faster than others. We can apply the LEMP algorithm in this model as well by adjusting the estimation of the expected reward based on the same global state for $K$ time units. Note that the assumptions in the analysis still hold, and we can still use Lezaud's result \cite{lezaud1998chernoff} in Lemma \ref{lemma:lezaud} (which bounds the probability of a large deviation from the stationary distribution) since it is applied to the local arms. Also, (\ref{eq:app12}) in Lemma $1$ (which bounds the estimation error of the global transition probabilities) holds as well. This adjustment would result in an additional $K$ constant term in the first term of the regret, as a consequences of (\ref{eq:condition2}). This additional constant does not violate the logarithmic order of the regret.

Before proceeding to prove Theorem \ref{th:regret}, we define the following auxiliary notation.
\begin{definition}
Let $T_1$ be the smallest integer, such that for all $t \geq T_1$ the following holds: $\overline{D}_s^i \leq \widehat{D}_s^i(t)$ for all $i \in \mathcal{N}, s \in \mathcal{S}$, and also $\widehat{D}_s^i(t) \leq \overline{D}_{s,max}^i$ for all $i\in\mathcal{K}_s, s \in \mathcal{S}$.
\end{definition}
The term $T_1$ captures the random time by which the exploration rates for all arms are sufficiently close to the desired exploration rates needed for achieving the desired logarithmic regret bound (as shown later).

\begin{IEEEproof}[Proof of Theorem \ref{th:regret}]
 The layout of the proof is as follows, first we show in Lemma \ref{lemma:T1}  that the expectation of the random time $T_1$ is bounded independently of $t$. Then, based on Lemma \ref{lemma:exp_regret_phi} we show in Lemma \ref{lemma:exploration} and Lemma \ref{lemma:exploitation},  that a logarithmic regret is obtained for all $t>T_1$, which yields the desired expected regret.

In the next Lemma we show that the expected value of $T_1$ is bounded under the LEMP algorithm. 
\begin{lemma}
\label{lemma:T1}
Assume that the LEMP algorithm is implemented as described in Section \ref{sec:Algorithm 1}. Then, $\mathbb{E}[T_1] < \infty$ is bounded independent of $t$. 
\end{lemma}

The proof of the Lemma \ref{lemma:T1} is given in Appendix \ref{append:proof_lemma1}.

The second step of the proof is to show that a logarithmic regret is obtained for all $t>T_1$, which yields the desired expected regret.
\begin{lemma}\label{lemma:exp_regret_phi}
Let $\tilde{T}^i(t) \triangleq \sum \limits_{n=1}^t \mathbbm{1}_{\{\phi(n)=i \neq i_n^*\}}$ denote the number of times arm $i$ was played when it was not the best arm during the $t$ first rounds.
Then the expected regret is upper bounded by: \vspace{-0.2cm}
\begin{equation}
\label{eq:main1}
\mathbb{E}_{\phi}[r(t)] \leq x_{\max}\mathbb{E}_{\phi}[T_1]+ x_{\max} \sum \limits_{i=1}^N \mathbb{E}_{\phi}[\tilde{T}^i(t)].
\end{equation}  \vspace*{-0.2cm}
\end{lemma}
We present the proof of this lemma in Appendix \ref{append:proof_lemma2}.

From (\ref{eq:main1}) we observe that it is sufficient to upper-bound the expected number of times an arm $i$ is played when this arm is sub-optimal. We will bound (\ref{eq:main1}) for the exploration and exploitation phases separately. Specifically, let $T^i_O(t)$, and $T^i_I(t)$, denote the time spent on sub-optimal arm $i$ in exploration and exploitation phases, respectively, by time $t$. Thus,
\begin{center}
$\tilde{T}^i(t)=T^i_O(t)+T^i_I(t)$.
\end{center}
The following two lemmas show that both $\mathbb{E}[T^i_O(t)]$ and $\mathbb{E}[T^i_I(t)]$ have a logarithmic order with time.

\begin{lemma}
\label{lemma:exploration}
The time spent by time $t$ in exploration phases for sub-optimal arm $i$ is bounded by: \vspace{0.3cm} \\
$\mathbb{E}[T^i_O(t)] \leq \vspace{0.3cm} \sum \limits_{i=1}^N 
\bigg[\frac{1}{3} [4(3A_i\cdot \log(t)+1)-1] \\
\vspace{0.3cm} \hspace{3cm}  +M^i_{\max} \cdot \log_4(3A_i\log(t)+1) \bigg]$. \vspace{-0.2cm}
\end{lemma}

\begin{lemma}
\label{lemma:exploitation}
The time spent by time $t$ in exploitation phases for sub-optimal arm $i$ is bounded by: 
\[
\mathbb{E}[T^i_I(t)] \leq 6 |\mathcal{S}| \cdot (\frac{|\mathcal{S}||\mathcal{X}_{\max}|}{\pi_{\min}}+2|\mathcal{S}|) \cdot \max_s \pi_s \cdot \lceil \log_4(\frac{3}{2}t+1) \rceil.
\]
\end{lemma}

The proofs of Lemma  \ref{lemma:exploration} and Lemma \ref{lemma:exploitation} are given in Appendix \ref{App.C} and Appendix \ref{App.D}, respectively.

Finally, we show at Appendix E, that combining the above four lemmas concludes the proof of Theorem \ref{th:regret}. 

\end{IEEEproof} \vspace{-0.2cm}
\section{Simulation Results}
\label{sec_simulation}
    \begin{table} [htbp!]
    \centering
    \begin{tabular}{ |p{1cm}||p{1.5cm}|p{1.5cm}|p{1.5cm}|}
    \hline
     & $N$ & $|\mathcal{S}|$ & $\Delta$\\
    \hline \hline
    S1 &  3  & 2 & 0.4\\
    \hline
    S2 &  6  & 2 & 0.4\\
    \hline
    S3 &  3  & 3 & 0.4\\
    \hline
    S4 &  3  & 2 & 0.2 \\
    \hline
    \end{tabular}  
    \caption{A description of the experiment parameters. We set $|\mathcal{X}_s^i| = 2$ in all scenarios. The values of all transition probabilities and state rewards under each scenario are described in the text.}
    \label{table:1}
    \end{table}

In this section we evaluate the regret of the LEMP algorithm numerically in four different scenarios.

We compare the LEMP algorithm to an extended version of the DSEE algorithm \cite{liu2013learning} 
which is an efficient and widely used algorithm in the RMAB settings, and to a strategy that chooses in the exploitation phases the best arm on average (i.e., competing against weak regret as in \cite{gafni2020learning,tekin2012online}). The DSEE algorithm uses deterministic sequencing of exploration and exploitation phases, however, it does not estimate the hardness parameter, and explores each arm $\frac{4L}{\Delta} \cdot \log(t)$ times, which results in oversampling bad arms to achieve the desired logarithmic regret. 
We simulate and compare the regret of these three algorithms averaged over $1000$ Monte-Carlo experiments, under four scenarios, denoted by S1, S2, S3, S4. In Table \ref{table:1} we summarize the number of arms, the number of global states, and the difference between the highest and the second highest values for each scenario. In all scenarios we set $L=6, I_L=0.1, I_G=0.1$.

\subsubsection{RMAB with exogenous process (S1)}
\begin{figure}[t!]
\centering \epsfig{file=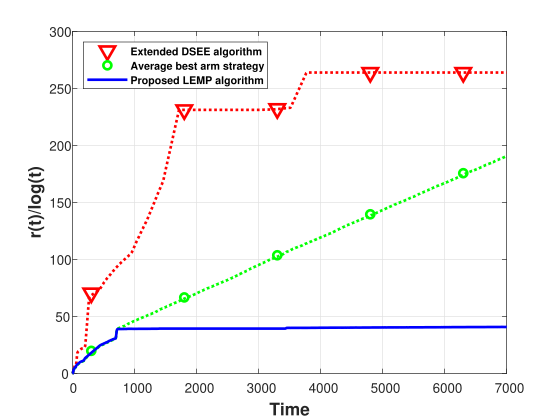,
width=0.45\textwidth}
\vspace{-0.2cm} 
\caption{S1: Performance comparison of the regret (normalized by log t).} \vspace*{-0.3cm} 
\label{fig:sim1}
\end{figure}

In Fig.~\ref{fig:sim1} we simulated S1 scenario. 
Here, the global state models the presence of the primary user that uses the entire bandwidth by a Gilbert-Eliot model \cite{liu2008link} that comprises a Markov chain with two binary states, where global state $s = 1$ denotes a transmitting primary user and $s = 0$ denotes a vacant channel, i.e., inactive primary user. 
To limit the interference to the primary user, a secondary user may choose to transmit over one of three possible channels (i.e., $N=3$), where the channels are modeled by a Finite-State Markovian Channel (FSMC), which is a tractable model widely used to capture the time-varying behavior of a radio communication channel (e.g., a Rayleigh fading channels \cite{wang1995finite}).
The transition probabilities of the global chain are $p_{00}=0.4$, $p_{10}=0.75$. 
In global state $1$, the local transition probabilities for all arms to transition from $1$ to $1$ and from $2$ to $1$, respectively, are: $p_{11} = [0.5,0.6,0.7]$, $p_{21} = [0.5,0.4,0.3]$, and the rewards for all arms at states $1,2$, respectively, are $r_1 = [4, 5.8, 1]$, $r_2 = [6, 8.2, 2]$.
In global state $2$, the local transition probabilities for all arms to transition from $1$ to $1$ and from $2$ to $1$, respectively, are: $p_{11} = [0.55,0.65,0.75]$, $p_{21} = [0.45,0.35,0.25]$, and the rewards for all arms at states $1,2$, respectively, are $r_1 = [10, 9, 2.5]$, $r_2 = [14,11,3]$.
It can be seen that LEMP significantly outperforms the extended DSEE algorithm. 
Fig.~\ref{fig:sim1} also shows the superior of LEMP against a strategy that chooses the best arm on average, demonstrating the gain in tracking the best arm at each step according to the global process evolution. 

\begin{figure}[t!]
\centering \epsfig{file=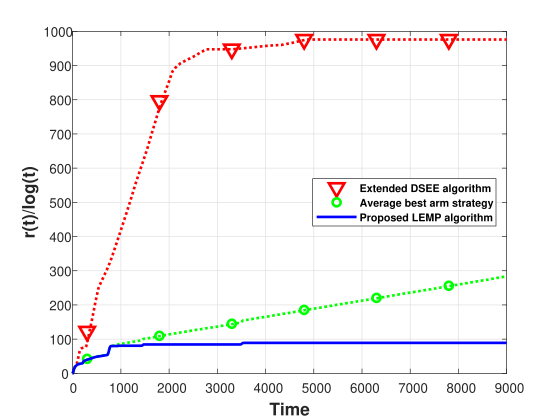,
width=0.45\textwidth}
\vspace{-0.2cm} 
\caption{S2: Performance comparison of the regret (normalized by log t).} \vspace{-0.4cm} 
\label{fig:sim_arms}
\end{figure}

\subsubsection{Increasing the number of arms (S2)} Next, we are interested in examine the regret in a larger system, i.e., we increased the number of arms to $6$ (S2). The results are depicted in Fig.~\ref{fig:sim_arms}. 
The transition probabilities of the global process is as in the previous scenario.
In global state $1$, the local transition probabilities for all arms to transition from $1$ to $1$ and from $2$ to $1$, respectively, are: $p_{11} = [0.5,0.6,0.7,0.7,0.6,0.5]$, $p_{21} = [0.5,0.4,0.3,0.3,0.4,0.5]$, and the rewards for all arms at states $1,2$, respectively, are $r_1 = [4, 5.8, 1,1.1,0.6,1.2]$, $r_2 = [6, 8.2, 2,1.9,0.9,2.2]$.
In global state $2$, the local transition probabilities for all arms to transition from $1$ to $1$ and from $2$ to $1$, respectively, are: $p_{11} = [0.55,0.65,0.75,0.75,0.65,0.55]$, $p_{21} = [0.45,0.35,0.25,0.25,0.35,0.45]$, and the rewards for all arms at states $1,2$, respectively, are $r_1 = [10, 9, 2.5,3,2.56,2.7]$, $r_2 = [14,11,3,2.8,3.1,3.3]$.
Increasing the number of arms is expected to decrease the performance under the extended DSEE algorithm, since more arms are sampled by the worst exploration rate. Indeed, it can be seen in Fig.~\ref{fig:sim_arms}, that the gap in the regret between LEMP and the extended DSEE algorithm is increased compared to the previous simulation. This is due to the fact that in the proposed LEMP algorithm, each arm is played according to its unique exploration rate (as a result of the online estimation of the hardness parameter), thus adding "bad" arms (i.e., arms with high exploration rate) does not significantly affect the LEMP performances. LEMP outperforms the strategy that chooses the best arm on average also in this scenario.

\subsubsection{Increasing the number of global states (S3)} 
\begin{figure}[t!]
\centering \epsfig{file=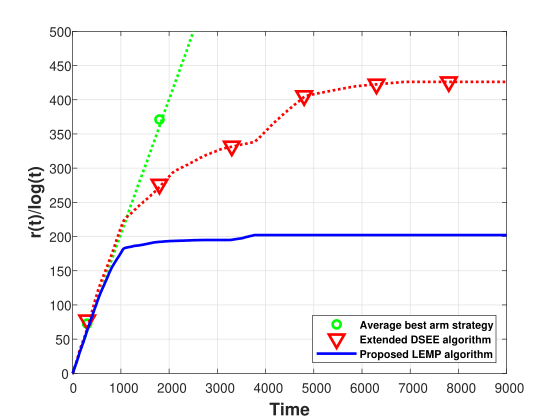,
width=0.45\textwidth}
\vspace{-0.2cm} 
\caption{S3: Performance comparison of the regret (normalized by log t).} \vspace{-0.3cm} 
\label{fig:sim_states}
\end{figure}

In Fig.~\ref{fig:sim_states} we increased the number of global states to $3$ (S3). 
The transition probabilities of the global chain are $p_{00}=0.85$, $p_{01}=0.1$, $p_{02}=0.05$, $p_{10}=0.08$, $p_{11}=0.85$, $p_{12}=0.07$, $p_{20}=0.06$, $p_{21}=0.09$, $p_{22}=0.85$.
In global state $1$, the local transition probabilities for all arms to transition from $1$ to $1$ and from $2$ to $1$, respectively, are: $p_{11} = [0.5,0.6,0.7]$, $p_{21} = [0.5,0.4,0.3]$, and the rewards for all arms at states $1,2$ are $r_1 = [4, 1, 1.2]$, $r_2 = [6,3,1.8]$, respectively.
In global state $2$, the local transition probabilities for all arms to transition from $1$ to $1$ and from $2$ to $1$, respectively, are: $p_{11} = [0.55,0.65,0.75]$, $p_{21} = [0.45,0.35,0.25]$, and the rewards for all arms at states $1,2$ are $r_1 = [5,9,4.5]$, $r_2 = [7,11,8.5]$, respectively.
In global state $3$, the local transition probabilities for all arms to transition from $1$ to $1$ and from $2$ to $1$, respectively, are: $p_{11} = [0.52,0.62,0.72]$, $p_{21} = [0.48,0.38,0.28]$, and the rewards for all arms at states $1,2$ are $r_1 = [9.9,9.5,14]$, $r_2 = [10.3,11.5,16]$, respectively.
In this scenario, for each global state $s$, there is a different best arm that is significantly better then the other two arms. 
Thus, playing the best arm on average in each time slot results in poor performances compared to LEMP, that tracks the best arm at each step, as can be seen in Fig.~\ref{fig:sim_states}. LEMP outperforms the extended DSEE algorithm also in this scenario.

\subsubsection{Decreasing the difference between the highest and the second highest values (S4)} 
\begin{figure}[t!]
\centering \epsfig{file=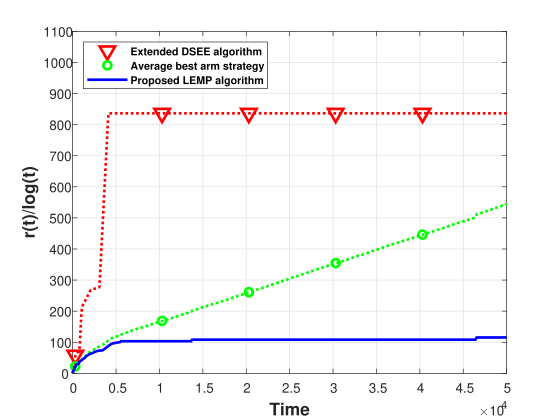,
width=0.45\textwidth}
\vspace{-0.2cm} 
\caption{S4: Performance comparison of the regret (normalized by log t).} \vspace{-0.4cm} 
\label{fig:Delta}
\end{figure}

Finally, we simulated S4, where we decreased the difference between the highest and the second highest values in global state $1$, compared to this difference in the first simulation (i.e., in Fig.~\ref{fig:sim1}). 
The transition probabilities of the global chain are $p_{00}=0.4$, $p_{10}=0.75$. 
In global state $1$, the local transition probabilities for all arms to transition from $1$ to $1$ and from $2$ to $1$, respectively, are: $p_{11} = [0.5,0.6,0.7]$, $p_{21} = [0.5,0.4,0.3]$, and the rewards for all arms at states $1,2$, respectively, are $r_1 = [4, 5.8, 1]$, $r_2 = [6, 9.2, 2]$.
In global state $2$, the local transition probabilities for all arms to transition from $1$ to $1$ and from $2$ to $1$, respectively, are: $p_{11} = [0.55,0.65,0.75]$, $p_{21} = [0.45,0.35,0.25]$, and the rewards for all arms at states $1,2$, respectively, are $r_1 = [10, 9, 2.5]$, $r_2 = [14,11,3]$.
Decreasing the difference between the highest and the second highest values in global state $1$ results in a high exploration rate used to distinguish between the two best arms in this state. As discussed in Section \ref{sec:Algorithm 1}, LEMP explores only these two arms using the high exploration rate, where the extended version of the DSEE algorithm explores all the arms with the high exploration rate. Indeed, as can be seen in Fig.~\ref{fig:Delta}, this effect results in a high regret under the extended DSEE algorithm as compared to LEMP, which also outperforms the strategy that chooses the best arm on average. 

We finally note that practically, it is well known that there is often a gap between the sufficient conditions required by theoretical analysis (often due to union-bounding events in the analysis) and practical conditions used for efficient online learning. While the sufficient conditions provided by the theoretical analysis in Section \ref{sec:regret} require to overestimate $\overline{D}^i_s(t)$ as in (\ref{eq:rate_estimate}), simulation results provide much better performance when higher values of $I_L$ and $I_G$ and lower value of $L$ are used. Thus, we tuned the parameters in all algorithms that we tested to achieve the best performance. \vspace{-0.2cm}

\section{Conclusion} 
\label{sec:conclusion}
We developed a novel Learning under Exogenous Markov Process (LEMP) algorithm for an extended version of the RMAB problem, where an exogenous Markov global process governs the distribution of the arms. Inspired by recent developments of sequencing methods of exploration and exploitation phases, LEMP estimates the hardness parameter of the problem which controls the size of exploration phases. During the exploitation phases, LEMP switches arms dynamically according to the global process evolution.
Simulation results support the theoretical analysis, and show superior performances of the proposed LEMP algorithm against competitive strategies. 
The model and analytical results presented in this paper lead to interesting future research directions. One direction would  consider the case where the global process is not directly observed (e.g., when the reward is a probabilistic function of the unobserved finite-state Markov process), by incorporating methods from hidden Markov model literature \cite{krishnamurthy2009partially}.
Another open problem is to derive a lower bound for the regret in our model, to understand the optimality properties of the LEMP algorithm. We leave this for future work.

\appendices{}

\section{Proof of Lemma \ref{lemma:T1}}\label{append:proof_lemma1}
\textit{Proof of Lemma \ref{lemma:T1}}:
First note that $\mathbb{E}[T_1]$ can be written as follows: \vspace{0.2cm}\\
$ \mathbb{E}[T_1]=\sum\limits_{n=1}^{\infty} n \cdot \mathbb{P}\left(T_1= n \right)=
\sum\limits_{n=1}^{\infty} \mathbb{P} \left(T_1\geq n \right)
\vspace{0.2cm} \\ \leq
\hspace{0.3cm} \sum\limits_{s\in\mathcal{S}} \sum\limits_{i\in\mathcal{K}_s}\sum\limits_{n=1}^{\infty} \sum\limits_{j=n}^{\infty}
\mathbb{P} \left(\widehat{D}_s^i(j)<\overline{D}_s^i \mbox{\;or\;}
\widehat{D}_s^i(j)> \overline{D}_{s,max}^i \right)\vspace{0.2cm} \\
\hspace{0.3cm} + \sum\limits_{s\in\mathcal{S}} \sum\limits_{i\nin\mathcal{K}_s}\sum\limits_{n=1}^{\infty} \sum\limits_{j=n}^{\infty}
\mathbb{P}\left(\widehat{D}_s^i(j)<\overline{D}_s^i \right). $\vspace{0.2cm}\\
Note that if we show that
 \begin{align}
 \label{eq:app1}
 \mathbb{P}\left(\widehat{D}_s^i(j)<\overline{D}_s^i \mbox{\;or\;}\widehat{D}_s^i(j)> \overline{D}_{s,max}^i \right) \leq C\cdot j^{-(2+\delta)},
\end{align}
for some constants $C>0, \delta > 0$ for all $i\in\mathcal{K}_s, s\in\mathcal{S}$ for all $j\geq n$, then we get: \vspace{0.2cm}
\\
$ \displaystyle \sum\limits_{s\in\mathcal{S}} \sum\limits_{i\in\mathcal{K}_s}\sum\limits_{n=1}^{\infty} \sum\limits_{j=n}^{\infty}
\mathbb{P}\left(\widehat{D}_s^i(j)<\overline{D}_s^i \mbox{\;or\;}\widehat{D}_s^i(j)> \overline{D}_{s,max}^i \right)
\vspace{0.2cm}\\\leq
\hspace{0.3cm} |\mathcal{S}| N C \left[\sum\limits_{j=1}^{\infty} j^{-(2+ \delta)}+\sum\limits_{n=2}^{\infty}\sum\limits_{j=n}^{\infty} j^{-(2+ \delta)}\right]
\vspace{0.2cm}\\\leq
\hspace{0.3cm} |\mathcal{S}| N C \left[\sum\limits_{j=1}^{\infty} j^{-(2+ \delta)}+\sum\limits_{n=2}^{\infty}\int\limits_{n-1}^{\infty} j^{-(2+ \delta)}dj\right]
\vspace{0.2cm}\\=
\hspace{0.3cm} |\mathcal{S}| N C \left[\sum\limits_{j=1}^{\infty} j^{-(2+ \delta)}+\frac{1}{1+\delta}\sum\limits_{n=2}^{\infty}(n-1)^{-(1+\delta)}\right] \\<\infty$,
\vspace{0.2cm}\\
which is bounded independent of $t$. Similarly, showing that $\mathbb{P}\left( \widehat{D}_s^i(j)<\overline{D}_s^i \right) \leq C\cdot j^{-(2+\delta)}$ for some constants $C, \delta > 0$ for all $i\nin\mathcal{K}_s,s\in\mathcal{S}$ for all $j\geq n$ completes the statement. 

\noindent\textbf{Step 1: Simplifying (\ref{eq:app1}):}
First, \vspace{0.2cm}\\
$
\mathbb{P}(\widehat{D}_s^i(t)<\overline{D}_s^i \quad or
\quad \widehat{D}_s^i(t)> \overline{D}_{s,max}^i)
\vspace{0.2cm}\\ = \mathbb{P} \bigg( \frac{4L}{\max\{\Delta,(\hat{V}_s^*(t)-\hat{V}_s^i(t))^2 - \epsilon\}} < \frac{4L}{(V_s^* - V_s^i)^2} \vspace{0.2cm} \\
\hspace*{0.5cm} \bigcup \frac{4L}{\max\{\Delta,(\hat{V}_s^*(t)-\hat{V}_s^i(t))^2 - \epsilon\}} > \frac{4L}{(V_s^* - V_s^i)^2 - 2\epsilon} \bigg)
\vspace{0.2cm} \\
\hspace{0.1cm}=
\mathbb{P}\bigg( \bigg[ \bigg((\hat{V}_s^*(t)- \hat{V}_s^i(t))^2- \epsilon> (V_s^*- V_s^i)^2  \vspace{0.2cm}\\ 
\hspace{1.7cm} \cap  (\hat{V}_s^*(t)- \hat{V}_s^i(t))^2-\epsilon \geq \Delta \bigg) \vspace{0.2cm}\\
 \hspace{1.2cm} \bigcup \bigg( \Delta>(V_s^*- V_s^i)^2 
\cap (\hat{V}_s^*(t)- \hat{V}_s^i(t))^2-\epsilon<\Delta
 \bigg) \bigg] \vspace{0.2cm}\\
\hspace{0.8cm} \displaystyle \bigcup \bigg[ \bigg( (\hat{V}_s^*(t)- \hat{V}_s^i(t))^2- \epsilon <
(V_s^*- V_s^i)^2 -2\epsilon \vspace{0.2cm}\\
\hspace{1.7cm} \cap (\hat{V}_s^*(t)- \hat{V}_s^i(t))^2-\epsilon \geq \Delta \bigg) \vspace{0.2cm}\\
 \hspace{1.2cm} \textstyle \bigcup \bigg( \Delta<(V_s^*- V_s^i)^2-2\epsilon 
 \cap (\hat{V}_s^*(t)- \hat{V}_s^i(t))^2-\epsilon<\Delta \bigg) \bigg] \bigg) \vspace{0.2cm}\\
\hspace{0.3cm} \leq \mathbb{P} \bigg ( \bigg [
(\hat{V}_s^*(t)- \hat{V}_s^i(t))^2- \epsilon> (V_s^*- V_s^i)^2  \vspace{0.2cm}\\
\hspace{1.7cm} \bigcup \Delta>(V_s^*- V_s^i)^2  \bigg ]\\
  \hspace{1.2cm} \displaystyle \bigcup \bigg[
 (\hat{V}_s^*(t)- \hat{V}_s^i(t))^2- \epsilon <
(V_s^*- V_s^i)^2 -2\epsilon \vspace{0.2cm}\\
  \hspace{1.7cm} \textstyle \bigcup (\hat{V}_s^*(t)- \hat{V}_s^i(t))^2-\epsilon<\Delta \bigg] \bigg).
$ \\
The probability for the second event on the RHS is zero, and the forth event lies inside the measure of the third event due to the fact that $i \in \mathcal{K}_s$.
Hence, 
\vspace{0.2cm}\\
$
\hspace{0.3cm} \mathbb{P}(\widehat{D}_s^i(t)<\overline{D}_s^i \quad or \quad \widehat{D}_s^i(t)> \overline{D}_{s,max}^i ) \vspace{0.2cm}\\
\hspace{-0.2cm} \leq \mathbb{P}(
(\hat{V}_s^*(t)- \hat{V}_s^i(t))^2-(V_s^*- V_s^i)^2 > \epsilon \vspace{0.2cm}\\
\hspace{0.7cm} \bigcup
(\hat{V}_s^*(t)- \hat{V}_s^i(t))^2- (V_s^*- V_s^i)^2 < -\epsilon ) \vspace{0.2cm}\\
\hspace{-0.2cm} = \mathbb{P}(|(\hat{V}_s^*(t)- \hat{V}_s^i(t))^2-
(V_s^*- V_s^i)^2| > \epsilon \} \vspace{0.2cm}\\
\hspace{-0.2cm} = \mathbb{P}(|(\hat{V}_s^*(t)- \hat{V}_s^i(t))^2 - (\hat{V}_s^*(t)- \hat{V}_s^i(t)) (V_s^*- V_s^i) \vspace{0.2cm}\\
\hspace{0.3cm} +(\hat{V}_s^*(t)- \hat{V}_s^i(t))(V_s^*- V_s^i)- (V_s^*- V_s^i)^2| > \epsilon \} \vspace{0.2cm}\\
\hspace{-0.2cm} = \mathbb{P}(|(\hat{V}_s^*(t)- \hat{V}_s^i(t)) [(\hat{V}_s^*(t)- \hat{V}_s^i(t))-(V_s^*- V_s^i))] \vspace{0.2cm}\\
\hspace{0.3cm} + (V_s^*- V_s^i)[(\hat{V}_s^*(t)- \hat{V}_s^i(t))-(V_s^*- V_s^i)]|> \epsilon \} $
 \begin{align}
 \label{eq:app2}
 \leq & \mathbb{P}(|(\hat{V}_s^*(t)- \hat{V}_s^i(t)) [(\hat{V}_s^*(t)- \hat{V}_s^i(t))-(V_s^*- V_s^i)]|> \frac{\epsilon}{2} )\nonumber \\
  + &\mathbb{P}(|(V_s^*- V_s^i)[(\hat{V}_s^*(t)- \hat{V}_s^i(t))-(V_s^*- V_s^i)]|> \frac{\epsilon}{2} ).
\end{align}

We continue by bounding the first term on the RHS of (\ref{eq:app2}). For every $R>0$, we have: 
\vspace{0.2cm}\\
$
\hspace{-0.1cm} \mathbb{P}(|(\hat{V}_s^*(t)- \hat{V}_s^i(t)) [(\hat{V}_s^*(t)- \hat{V}_s^i(t))-(V_s^*- V_s^i)]|> \frac{\epsilon}{2} ) \vspace{0.2cm}\\ \leq
\hspace{0.0cm} \mathbb{P}(|(\hat{V}_s^*(t)- \hat{V}_s^i(t))-(V_s^*- V_s^i)|>1)\vspace{0.2cm}\\
\hspace{0.6cm} +\mathbb{P}(|(\hat{V}_s^*(t)- \hat{V}_s^i(t))-(V_s^*- V_s^i)|> \frac{\epsilon}{2(R+1)}) \vspace{0.2cm}\\
\hspace{0.6cm} +\mathbb{P}(|(V_s^*- V_s^i)+1|>R)
 \vspace{0.2cm}\\ \leq
\hspace{0.1cm} 2 \mathbb{P}(|(\hat{V}_s^*(t)- \hat{V}_s^i(t))-(V_s^*- V_s^i)|> \frac{\epsilon}{2(R+1)}) \vspace{0.2cm}\\
\hspace{0.1cm} +\mathbb{P}(V_s^*+1>R ). \vspace{0.2cm}\\
$
We choose $R=V_s^*+1$. Then, the second term is equal to 0. 
We proceed with the first term:\vspace{0.2cm}\\
$
\hspace{0.1cm} 2\cdot \mathbb{P}(|(\hat{V}_s^*(t)- \hat{V}_s^i(t))-(V_s^*- V_s^i)|> \frac{\epsilon}{2(V_s^*+2)}) $
\begin{align}
\label{eq:app3}
\hspace{-2.7cm} \leq  2\mathbb{P}(|\hat{V}_s^*(t)-V_s^*|
>\frac{\epsilon}{4(V_s^*+2)} )\nonumber \vspace{0.2cm}\\
\hspace{-1.2cm} +2\mathbb{P}(|\hat{V}_s^i(t)-V_s^i)|>\frac{\epsilon}{4(V_s^*+2)}).
\end{align}

\hspace{-0.4cm} We next bound the second term on the RHS of (\ref{eq:app2}). For every $R'>0$, we have: 
\vspace{0.2cm}\\
$
\mathbb{P}(|(V_s^*- V_s^i)[(\hat{V}_s^*(t)- \hat{V}_s^i(t))-(V_s^*- V_s^i)]|> \frac{\epsilon}{2} ) \vspace{0.2cm}\\ \leq
\mathbb{P}(V_s^*>R') \vspace{0.2cm}\\
+\mathbb{P}(|(\hat{V}_s^*(t)- \hat{V}_s^i(t))-(V_s^*- V_s^i)|>\frac{\epsilon}{2(R'+1)} ).\vspace{0.2cm}\\ $
We now choose $R'=R=V_s^*+1$, so the first term is equal to 0. 
We continue with the second term:\vspace{0.2cm}\\
$
\mathbb{P}(|(\hat{V}_s^*(t)- \hat{V}_s^i(t))-(V_s^*- V_s^i)|>\frac{\epsilon}{2(R'+1)}) \leq$
\begin{align}
\label{eq:app4}
\hspace{-3.4cm} \mathbb{P}(|\hat{V}_s^*(t)-V_s^*|
>\frac{\epsilon}{4(V_s^*+2)}) \vspace{0.2cm}\\
\hspace{-2.4cm}+ \nonumber \mathbb{P}(|\hat{V}_s^i(t)-V_s^i)|>\frac{\epsilon}{4(V_s^*+2)}).
\end{align}

\hspace{-0.4cm} By combining (\ref{eq:app3}) and (\ref{eq:app4}) we get:\vspace{0.2cm}\\ 
$
\mathbb{P}(\overline{D}_s^i(t)<\widehat{D}_s^i \quad or
\quad \overline{D}_s^i(t)> \overline{D}_{s,max}^i) $
\begin{align}
\label{eq:app5}
\hspace{-0cm} \leq 6\cdot \max \left\{ \mathbb{P} \left(|\hat{V}_s^*(t)-V_s^*|
>\frac{\epsilon}{4(V_s^*+2)} \right) , \nonumber \right.\vspace{0.2cm}\\\left.
\hspace{-2cm} \mathbb{P} \left( |\hat{V}_s^i(t)-V_s^i|>\frac{\epsilon}{4(V_s^*+2)} \right) \right\}.
\end{align}

\noindent\textbf{Step 2: Bounding (\ref{eq:app5}):}
We first bound the second term in (\ref{eq:app5}) (the first term is bounded similarly). \vspace{0.2cm}\\ \hspace{0.2cm} 
$\mathbb{P} \left( |\hat{V}_s^i(t)-V_s^i|>\frac{\epsilon}{4(V_s^*+2)} \right) \vspace{0.2cm}\\ =
\hspace{0.2cm} \mathbb{P} \left( |\sum \limits_{s'\in\mathcal{S}}\hat{p}_{ss'}(t)\hat{\mu}_{s'}^i(t)-\sum \limits_{s'\in\mathcal{S}}p_{ss'}\mu_{s'}^i|>\frac{\epsilon}{4(V_s^*+2)} \right)
\vspace{0.2cm}\\ \leq \hspace{0.2cm}
\mathbb{P} \left(|\sum \limits_{s'\in\mathcal{S}}(\hat{p}_{ss'}(t)\hat{\mu}_{s'}^i(t)-p_{ss'}\mu_{s'}^i-\frac{\epsilon}{4(V_s^*+2)|\mathcal{S}|} )| > 0 \right) \vspace{0.2cm}\\ \leq \hspace{0.2cm}
\sum \limits_{s'\in\mathcal{S}} \mathbb{P} \left(|\hat{p}_{ss'}(t)\hat{\mu}_{s'}^i(t)-p_{ss'}\mu_{s'}^i| > \frac{\epsilon}{4(V_s^*+2)|\mathcal{S}|} \right). \vspace{0.2cm}\\
$ 
Following similar steps as we did to obtain (\ref{eq:app5}) from (\ref{eq:app2}), we get \vspace{0.2cm}\\ 
$\mathbb{P} \left(|\hat{p}_{ss'}(t)\hat{\mu}_{s'}^i(t)-p_{ss'}\mu_{s'}^i| > \frac{\epsilon}{4(V_s^*+2)|\mathcal{S}|} \right)$ \\
\begin{align}
\label{eq:app10}
 \leq \vspace{0.2cm}
2 \mathbb{P}(|\hat{p}_{ss'}(t)-p_{ss'}|> \frac{\epsilon}{16(V_s^*+2)(x_{\max}+2)|\mathcal{S}|}) \\ 
\label{eq:app11}
+ \vspace{0.2cm} \mathbb{P}(|\hat{\mu}_{s'}^i(t)-\mu_{s'}^i|> \frac{\epsilon}{16(V_s^*+2)(x_{\max}+2)|\mathcal{S}|}).
\end{align}
To complete the statement, we need to bound (\ref{eq:app10}) and (\ref{eq:app11}).
To bound (\ref{eq:app11}), we will use Lezaud's result \cite{lezaud1998chernoff}. 
\begin{lemma} \label{lemma:lezaud} \cite{lezaud1998chernoff} Consider a finite-state, irreducible Markov chain $\{X_t\}_{t \geq 1}$ with state space $S$, matrix of transition probabilities $P$, an initial distribution $q$,  and stationary distribution $\pi$. Let
$N_\textbf{q}=\left \| (\frac{q_x}{\pi_x}, x \in S) \right \|_2 $.
Let $\widehat{P}=P'P $ be the multiplicative symmetrization of $P$ where $P'$ is the adjoint of $P$ on $l_2(\pi)$. Let $\epsilon= 1-\lambda_2$, where $\lambda_2$ is the second largest eigenvalue of the matrix $P'$. $\epsilon$ will be referred to as the eigenvalue gap of $P'$. Let $f:S \rightarrow \mathcal{R}$ be such that $\sum\limits_{y \in S} \pi_yf(y)=0, \quad \|f\|_2 \leq 1$ and $0 \leq \|f\|_2^2 \leq 1$ if $P'$ is irreducible. Then, for any positive integer $n$ and all $0<\lambda\leq 1$, we have:\vspace{0.2cm}
$$Pr \displaystyle \left(\frac{1}{n} \sum\limits_{t=1} ^{n}f(X_t) \geq \lambda      \right) \leq N_\textbf{q} \exp [-\frac{n \lambda^2 \epsilon}{12}].$$
\end{lemma}
Consider an initial distribution $\textbf{q}_s^i$ for the $i$th arm in global state $s$. We have:
\begin{center}
$ \displaystyle \left \| (\frac{q_s^i(x)}{\pi_s^i(x)}, x \in \mathcal{X}_s^i) \right \|_2 \leq
  \sum\limits_{x \in \mathcal{X}_s^i} \left \| \frac{q_s^i(x)}{\pi_s^i(x)} \right \|_2 \leq \frac{1}{\pi_{min}}$.
\end{center}

Before applying Lezaud's bound, we pay attention for the following:
(i) The sample means $\{\hat{\mu}_s^i(t)\}$ are calculated only from measurements in the set $\mathcal{V}_i$. As discussed in Section \ref{ssec:exploration}, these measurements are equivalent to a sample path generated by continuously sampling the Markov chain. Hence, we can apply Lezaud's bound to upper bound (\ref{eq:app11}).
(ii) By the construction of the algorithm, (\ref{eq:condition1}) ensures that once exploitation phases are executed (which are deterministic), the event $T_s^i(t)\geq\frac{(2+\delta)}{\epsilon^2 I_L}\log(t)$ for $\delta>0$ arbitrarily small surely occurs\footnote{We point out that a precise statement requires to set $(2+2\delta)$ in (\ref{eq:condition1}) and the statement holds for all $t>D$, where $D$ is a finite deterministic value. However, since $\delta>0$ is arbitrarily small and is not a design parameter, we do not present it explicitly when describing the algorithm to simplify the presentation.}. During exploration phases, the randomness of SB1 (say for arm $r\neq i$) affects $T_s^i(t)$ since SB1 can be very long (with small probability) and then $T_s^i(t)\geq\frac{(2+\delta)}{\epsilon^2 I_L}\log(t)$ might not hold until the end of the phase once the algorithm corrects the exploration gap by condition (\ref{eq:condition1}). Therefore, we define $E_i(t)$ as the event when all SB1 phases that have been executed by time $t$ are smaller than $\delta\cdot t$. When event $E_i(t)$ occurs we have $T_s^i(t)\geq\frac{(2+\delta)}{\epsilon^2 I_L}\log(t)$ (for all $t>D$, for a sufficiently large finite deterministic value $D$). Then, for all $i$ and $s'$, we have: 
\vspace{0.2cm}\\
$
\mathbb{P}(|\hat{\mu}_{s'}^i(t)-\mu_{s'}^i|> \frac{\epsilon}{16(V_s^*+2)(x_{\max}+2)|\mathcal{S}|})\vspace{0.2cm}\\=
\mathbb{P}(|\hat{\mu}_{s'}^i(t)-\mu_{s'}^i|> \frac{\epsilon}{16(V_s^*+2)(x_{\max}+2)|\mathcal{S}|}, \mbox{\;$E_i(t)$ occurs})\vspace{0.2cm}\\+
\mathbb{P}(|\hat{\mu}_{s'}^i(t)-\mu_{s'}^i|> \frac{\epsilon}{16(V_s^*+2)(x_{\max}+2)|\mathcal{S}|} \\\hspace{0.6cm}, \mbox{\;$E_i(t)$ does not occur})$
\begin{align}
\nonumber
 \hspace*{-0.2cm} \leq \mathbb{P}\bigg(|\hat{\mu}_{s'}^i(t)-\mu_{s'}^i|> &\frac{\epsilon}{16(V_s^*+2)(x_{\max}+2)|\mathcal{S}|} \\\label{eq:app6}
 , &\mbox{\;$E_i(t)$ occurs}\bigg)\vspace{0.2cm}\\
 \label{eq:app7} \hspace{0.5cm}+\mathbb{P}(\mbox{\;$E_i(t)$ does not occur}).
\end{align}

We next bound (\ref{eq:app11}) by bounding (\ref{eq:app6}) and (\ref{eq:app7}): 
\vspace{0.2cm}\\
We define $O^{i,x}_s(t)$ as the number of occurrences of local state $x$ on arm $i$ in global state $s$ up to time t, and we first look at: 
\vspace{0.2cm}\\
$
\mathbb{P}(\hat{\mu}_{s'}^i(t)-\mu_{s'}^i> \frac{\epsilon}{16(V_s^*+2)(x_{\max}+2)|\mathcal{S}|}, E_i(t)) \vspace{0.2cm}\\ =
\mathbb{P}\bigg(\sum\limits_{x \in \mathcal{X}_s^i} x \cdot O^{i,x}_s(t)-T_s^i(t) \sum\limits_{x \in \mathcal{X}_s^i} x \cdot \pi_s^i(x) \vspace{0.2cm}\\ \hspace*{1cm} > \frac{T_s^i(t)\cdot \epsilon}{16(V_s^*+2)(x_{\max}+2)|\mathcal{S}|}, E_i(t) \bigg) \vspace{0.2cm}\\ \leq
\sum\limits_{x \in \mathcal{X}_s^i} \mathbb{P}\bigg(\frac{\sum \limits_{n=1}^t \textbf{1} (x_s^i(n)=x)-T_s^i(t) \pi_s^i(x) } {\hat{\pi}_s^i(x) \cdot T_s^i(t)} \vspace{0.2cm}\\  \hspace*{1cm}> \frac{T_s^i(t)\cdot \epsilon}{16(V_s^*+2)(x_{\max}+2)|\mathcal{S}| |\mathcal{X}_s^i| \cdot x \hat{\pi}_s^i(x) }, E_i(t)\bigg) \vspace{0.2cm}\\ \leq
|\mathcal{X}_s^i|  N_{s,\textbf{q}}^{(i)}$ \vspace{0.2cm}\\
$\cdot \exp (-T_s^i(t)  \frac{\epsilon^2}{(16(V_s^*+2)(x_{\max}+2)|\mathcal{S}|)^2 \cdot x_{\max}^2  |\mathcal{X}_s^i|^2 \hat{\pi}_{\max}^2}  \frac{\bar{\lambda}_{\min}}{12}) \vspace{0.2cm}\\ $
and due to $E_i(t)$: $T_s^i(t)> \frac{2+\delta}{\epsilon^2 I_L} \cdot log(t)$, so we have: 
\vspace{0.2cm}\\
$
\mathbb{P}(\hat{\mu}_{s'}^i(t)-\mu_{s'}^i > \frac{\epsilon}{16(V_s^*+2)(x_{\max}+2)|\mathcal{S}|}, E_i(t)) \vspace{0.2cm}\\ \leq
\hspace{0.3cm}\frac{|\mathcal{X}_{max}|}{\pi_{min}}$ \vspace{0.2cm}\\
$\exp(- \frac{(2+ \delta)}{\epsilon^2 I_L} \frac{\epsilon^2 \cdot \bar{\lambda}_{\min}}{12 \cdot 16^2(V_s^*+2)^2 (x_{\max}+2)^2  x_{\max}^2 |\mathcal{X}_s^i|^2  |\mathcal{S}|^2 \hat{\pi}_{\max}^2} \log(t))\vspace{0.2cm}$
\begin{equation}
\leq
\hspace{0.3cm} \frac{|\mathcal{X}_{max}|}{\pi_{min}} \cdot e^{-(2+\delta)\cdot \log(t)}=\frac{|\mathcal{X}_{max}|}{\pi_{min}} \cdot t^{-(2+\delta)},
\end{equation}
for some $\delta>0$ arbitrarily small. \vspace{0.2cm}\\
Together with applying Lemma 3 to $-f$, we get the bound for (\ref{eq:app6}).
We next upper bound (\ref{eq:app7}). When event $E_i(t)$ does not occur, there exists an SB1 phase (i.e., hitting time) which is greater than $\delta\cdot t$. 
Since all arms are finite state and irreducible Markov chains (i.e., the hitting time of some state $x$ starting from state $y$ is almost surely finite), we have that $\mathbb{P}(\mbox{\;$E_i(t)$ does not occur})\leq e^1 \cdot e^{-\frac{\delta}{|\mathbf{X}_{\max}|} t}$.
Therefore for all $t>D$, for a sufficiently large finite deterministic value $D$, we have  $\mathbb{P}(\mbox{\;$E_i(t)$ does not occur}) \leq \frac{|\mathcal{X}_{\max}|}{\pi_{\min}}t^{-(2+\delta)}$,  
which completes (\ref{eq:app11}). \vspace{0.2cm}\\
(\ref{eq:app10}) is bounded by: 
\begin{align}
\label{eq:app12}
 \displaystyle \mathbb{P}\left(\hat{p}_{ss'}(t)-p_{ss'}>\frac{\epsilon}{16(V_s^*+2)(x_{\max}+2)|\mathcal{S}|}\right) \nonumber\vspace{0.2cm}\\  \leq \exp\left(-2N_s(t)\cdot\left(\frac{\epsilon}{16(V_s^*+2)(x_{\max}+2)|\mathcal{S}|}\right)^2\right) \end{align}
 
The bound in (\ref{eq:app12}) follows similar steps as in Part A of Appendix A in \cite{yemini2021restless}. 
Using condition (\ref{eq:condition2}),$N_s(t) > \frac{2+\delta}{\epsilon^2 \cdot I_G} \cdot \log(t)$ for some arbitrarily small $\delta>0$ , we conclude that (\ref{eq:app12}) is bounded by $t^{(2+\delta)}$, which proves that (\ref{eq:app1}) is bounded by 
$4 \frac{|\mathcal{X}_{max}|}{\pi_{min}} \cdot t^{-(2+\delta)}$.

Finally, showing that $\mathbb{P}(\widehat{D}_s^i(j)<\overline{D}_s^i)\leq j^{-(2+\delta)}$ for some $\delta > 0$ for all $i\nin\mathcal{K}_s$ for all $j\geq n$ follows similar steps as showed by handling $\widehat{D}_s^i(j)<\overline{D}_s^i$ when proving (\ref{eq:app1}). 
Thus, Lemma \ref{lemma:T1} follows. \vspace{-0.3cm}

\section{Proof of Lemma \ref{lemma:exp_regret_phi}} 
\label{append:proof_lemma2}

Next, we prove the upper bound (\ref{eq:main1}). 
Note that the regret can be written as follows: \\
$ \mathbb{E}_{\phi}[r(t)]= \mathbb{E}_{\phi}\bigg[\sum \limits_{n=1}^{t} \sum \limits_{i\in V_{\text{e}}(n)} (x^{i^*_n}_{s_n}(n) - x^i_{s_n}(n))\mathbbm{1}_{\{\phi(n)=i\}}\bigg]$ 
\begin{align}
\label{eq:app17}
 & \hspace{-0.6cm} =\mathbb{E}_{\phi}\bigg[\sum \limits_{n=1}^{T_1} \sum \limits_{i\in V_{\text{e}}(n)} (x^{i^*_n}_{s_n}(n) - x^i_{s_n}(n))\mathbbm{1}_{\{\phi(n)=i\}}\bigg] \vspace{0.2cm}\\\label{eq:app18}
 +& \mathbb{E}_{\phi}\bigg[\sum \limits_{n=T_1+1}^{t} \sum \limits_{i\in V_{\text{e}}(n)} (x^{i^*_n}_{s_n}(n) - x^i_{s_n}(n))\mathbbm{1}_{\{\phi(n)=i\}}\bigg]. 
\end{align}
By applying Lemma \ref{lemma:T1}, we obtain that (\ref{eq:app17}) is bounded independent of $t$:
\[\mathbb{E}_{\phi}\bigg[\sum \limits_{n=1}^{T_1} \sum \limits_{i\in V_{\text{e}}(n)} (x^{i^*_n}_{s_n}(n) - x^i_{s_n}(n))\mathbbm{1}_{\{\phi(n)=i\}}\bigg]  \leq x_{\max}\mathbb{E}_{\phi}[T_1],\] 
which results in the additional constant term $O(1)$ in the regret bound in (\ref{eq:regret}) which is independent of $t$.

Next, we upper bound (\ref{eq:app18}). Note that for all $t>T_1$, we have: 
\begin{align}
\label{eq:app20}
 \overline{D}_s^i \leq \widehat{D}_s^i(t) \leq \overline{D}_{s,max}^i,
\end{align} 
for all $s\in\mathcal{S}, i\in\mathcal{K}_s$, and we have the LHS of the inequality for $i\nin\mathcal{K}_s$.
For convenience, we will develop (\ref{eq:app18}) between $n=1$ and $t$ with (\ref{eq:app20}) (and the LHS for $i\nin\mathcal{K}_s$) holds for all $1\leq n \leq t$, which upper bounds (\ref{eq:app18}) between $n=T_1+1$ and $t$:\\
$\vspace{0.3cm} 
\mathbb{E}_{\phi}\bigg[\sum \limits_{n=T_1+1}^{t} \sum \limits_{i\in V_{\text{e}}(n)} (x^{i^*_n}_{s_n}(n) - x^i_{s_n}(n))\mathbbm{1}_{\{\phi(n)=i\}}\bigg]$ \\
\begin{equation}
\label{eq:app21}
\leq \mathbb{E}_{\phi}\bigg[\sum \limits_{n=1}^{t} \sum \limits_{i\in V_{\text{e}}(n)} (x^{i^*_n}_{s_n}(n) - x^i_{s_n}(n))\mathbbm{1}_{\{\phi(n)=i\}}\bigg].
\end{equation}

Finally, note that:\vspace{0.3cm}\\
$\vspace{0.1cm} \mathbb{E}_{\phi}\bigg[\sum \limits_{n=1}^{t} \sum \limits_{i\in V_{\text{e}}(n)} (x^{i^*_n}_{s_n}(n) - x^i_{s_n}(n))\mathbbm{1}_{\{\phi(n)=i\}}\bigg]$
\begin{equation}
\label{eq:app22}
\leq x_{\max} \sum \limits_{i=1}^N \mathbb{E}_{\phi}[\tilde{T}^i(t)].
\end{equation}

\section{Proof of Lemma \ref{lemma:exploration}}\label{App.C}
\textit{Proof of Lemma \ref{lemma:exploration}}:
We first upper bound the number of exploration phases $n_O^i(t)$ for each arm (say $i$) by time $t$. If the player has started the $n^{th}$ exploration phase, we have by (\ref{eq:condition1}) and the fact that $t\geq T_1$:
\[\sum \limits_{n=1}^{n_O^i(t)} 4^{n-1} =\frac{1}{3}(4^{n_O^i(t)}-1) \leq A_i \cdot \log (t) .\]
Hence, $n_O^i(t) \leq \lfloor \log_4(3A_i\log(t)+1) \rfloor +1$.

Next, note that exploration phase $n_O^i(t)$ for arm $i$ consists of the time until the last state observed at the $(n_O^i(t)-1)^{th}$ exploration phase $\gamma^i(n_O^i-1)$ is observed again (i.e., SB1 sub-block), and another $4^{n_O^i(t)}$ time slots.
Thus, the time spent by time $t$ in exploration phases for arm $i$ is bounded by: \vspace{0.3cm} \\
$\vspace{0.3cm}\mathbb{E}[T_O^i(t)] \leq \sum \limits_{n=0}^{n_O^i(t)-1}(4^n+M^i_{\max})= \\
\vspace{0.3cm} \hspace{0.3cm}\frac{1}{3}(4^{n_O^i(t)}-1)+ M^i_{\max} \cdot n_O^i(t) \leq \\
\vspace{0.3cm} \hspace{0.3cm} \frac{1}{3} [4(3A_i\cdot \log(t)+1)-1]+
M^i_{\max} \cdot \log_4(3A_i\log(t)+1)$.

\section{Proof of Lemma \ref{lemma:exploitation}}\label{App.D}
\textit{Proof of Lemma \ref{lemma:exploitation}}:
We first upper bound the number of exploitation phases by time $t$, $n_I(t)$.
By time $t$, at most $t$ time slots have been spent on exploitation phases. Thus, we have:
\[\sum \limits_{n=1}^{n_I(t)} 2\cdot 4^{n-1} \leq t,\]
which implies that $\frac{2}{3}(4^{n_I(t)}-1) \leq t$.
Hence,
\beq
\label{eq:num_exploit_phases}
n_I(t) \leq \lceil \log_4(\frac{3}{2}t+1) \rceil.
\eeq
Next, we use (\ref{eq:num_exploit_phases}) to bound the regret caused by choosing sub-optimal arms in exploitation phases. 
Let $T^i_{s,I}(t)$ denotes the time spent on sub-optimal arm $i$ in global state $s$ in exploitation phases, by time $t$ (note that $T^i_I(t) = \sum \limits_{s \in \mathcal{S}}T^i_{s,I}(t) \leq |\mathcal{S}| \max_s \{T^i_{s,I}(t) \} $).
We define $Pr[i,s,n]$ as the probability that a sub-optimal arm $i$ is played when the global state is $s$ in the $n^{th}$ exploitation phase. From (\ref{eq:num_exploit_phases}) we have:
\begin{align}
\mathbb{E}[T^i_{s,I}(t)] &\leq \sum \limits_{n=1}^{n_I} 2\cdot 4^{n-1} \cdot \pi_s \cdot Pr[i,s,n] \nonumber\\
&\leq \sum \limits_{n=1}^{\lceil \log_4(\frac{3}{2}t+1) \rceil} 2\cdot 4^{n-1} \cdot \pi_s \cdot Pr[i,s,n]\nonumber\\
&\leq\sum \limits_{n=1}^{\lceil \log_4(\frac{3}{2}t+1) \rceil} 3t_n \cdot \pi_s \cdot Pr[i,s,n],\label{eq:time_bad_exploit}
\end{align} 
where $t_n$ denotes the starting time of the $n^{th}$ exploitation phase and (\ref{eq:time_bad_exploit}) follows from the fact that $t_n \geq \frac{2}{3} 4^{n-1} $.
Note that it suffices to show that $Pr[i,s,n]$ has an order of $t_n^{-1}$ so as to obtain a logarithmic order with time for the summation in (\ref{eq:time_bad_exploit}).

Next, we bound $Pr[i,s,n]$. We define $C_{s,t}^i= \sqrt{L \log(t)/T_s^i(t)}$,$C_{s,t}^*= \sqrt{L \log(t)/T_s^*(t) }$ , where $T_s^*(t)$ denotes the number of plays on the best arm of global state $s$, $i^*_s$, by time $t$.
\beq
\bea{l}
\label{eq:app13}
Pr[i,s,n]= \mathbb{P}(\hat{V}_s^i(t_n) \geq \hat{V}_s^*(t_n) )\vspace{0.2cm}\\ \hspace{0.5cm}
\leq \mathbb{P}(\hat{V}_s^*(t_n) \leq V_s^*- C_{s,t_n}^* ) \vspace{0.2cm}\\ \hspace{1cm}
+\mathbb{P}(\hat{V}_s^i(t_n) \geq V_s^i+ C_{s,t_n}^i) \vspace{0.2cm}\\ \hspace{1cm}
+\mathbb{P}(V_s^* <V_s^i+ C_{s,t_n}^i+ C_{s,t_n}^*).
\ena
\eeq
We first show that the third term in (\ref{eq:app13}) is zero. Note that from (\ref{eq:condition1}) we have:
\begin{center}
$\vspace{0.3cm} T_s^i(t)> \max \{\widehat{D}_s^i(t),\frac{2}{\epsilon^2 I_L} \} \cdot \log t_n$,
\end{center}
and from (\ref{eq:app20}) and the fact that $\overline{D}_s^i \leq \max_i \overline{D}_s^i $, we have:
\begin{center}
$\min\left\{T_s^* , T_s^i\right\} \geq \overline{D}_s^i \cdot \log t_n$.\vspace{0.2cm}
\end{center}
As a result,
\begin{center}
\label{eq:app14}
$\bea{l}
\mathbb{P}(V_s^* <V_s^i+ C_{s,t_n}^i+ C_{s,t_n}^*)\vspace{0.2cm}\\
=\mathbb{P}( V_s^* - V_s^i < \sqrt{\frac{L \log t_n}{T_s^i(t_n)}}+\sqrt{\frac{L \log t_n}{T_s^*(t_n)}})\vspace{0.2cm}\\
\leq \mathbb{P}( V_s^* - V_s^i < 2\sqrt{\frac{L \log t_n}{\min\left\{T_s^*(t_n) , T_s^i(t_n)\right\}}})\vspace{0.2cm}\\
=\mathbb{P}( (V_s^* - V_s^i)^2< \frac{4L \log t_n}{\min\left\{T_s^*(t_n) , T_s^i(t_n)\right\}}) \vspace{0.2cm}\\
= \mathbb{P}(\min\left\{T_s^*(t_n) , T_s^i(t_n)\right\} < \overline{D}_s^i \cdot \log t_n )=0.
\ena$
\end{center}
Therefore, we can rewrite (\ref{eq:app13}) as follows:
\beq
\bea{l}
\label{eq:app15}
Pr[i,s,n] \leq \mathbb{P}(\hat{V}_s^i(t_n) \geq \hat{V}_s^*(t_n) )\vspace{0.2cm}\\ \hspace{0.5cm}
\leq \mathbb{P}(\hat{V}_s^*(t_n) \leq V_s^*- C_{s,t_n}^* ) \vspace{0.2cm}\\ \hspace{1cm}
+\mathbb{P}(\hat{V}_s^i(t_n) \geq V_s^i+ C_{s,t_n}^i). \vspace{0.2cm} \hspace{1cm}
\ena
\eeq
Next, we bound both terms on the RHS of (\ref{eq:app15}).
Using similar steps as we used for bounding the second term in (\ref{eq:app5}), we get:
\beq
\bea{l}
\label{eq:app16}
\vspace{0.2cm} \mathbb{P}(\hat{V}_s^i(t_n) - V_s^i \geq C_{s,t_n}^i) \\ \vspace{0.2cm}
\leq 2|\mathcal{S}|\mathbb{P}(|\hat{p}_{ss'}(t_n)-p_{ss'}| \geq \frac{1}{4|\mathcal{S}| (x_{\max}+2)}\cdot C_{s,t_n}^i) \\
+ |\mathcal{S}|\mathbb{P}(|\hat{\mu}_{s'}^i(t_n)-\mu_{s'}^i| \geq \frac{1}{4|\mathcal{S}| (x_{\max}+2)}\cdot C_{s,t_n}^i).
\ena
\eeq
The second term in (\ref{eq:app16}) is bounded similarly as in (\ref{eq:app11}): \vspace{0.2cm}\\
$|\mathcal{S}|\mathbb{P}(|\hat{\mu}_{s'}^i(t_n)-\mu_{s'}^i| \geq \frac{1}{4|\mathcal{S}| (x_{\max}+2)}\cdot C_{s,t_n}^i) \vspace{0.2cm}\\
 \leq \frac{|\mathcal{S}||\mathcal{X}_{\max}|}{\pi_{\min}} \vspace{0.2cm}\\ \cdot \exp(-T_s^i(t_n) \frac{L \log(t_n)}{T_s^i(t_n)}\frac{1}{16(x_{\max}+2)^2|\mathcal{S}|^2} \frac{\bar{\lambda}_{\min}}{12  (x_{max} |\mathcal{X}_{\max}|  \pi_{\max})^2}) \vspace{0.2cm}\\ \leq
\frac{|\mathcal{S}||\mathcal{X}_{\max}|}{\pi_{\min}} \cdot t_n^{-1}, \vspace{0.2cm}\\
$
where the last inequality is due to (\ref{eq:L}).
The first term in (\ref{eq:app16}) is bounded similarly as in (\ref{eq:app10}): \vspace{0.2cm}\\
$
2|\mathcal{S}|\mathbb{P}(|\hat{p}_{ss'}(t_n)-p_{ss'}| \geq \frac{1}{4|\mathcal{S}|(x_{\max}+2)}\cdot C_{s,t_n}^i) \vspace{0.2cm}\\ \leq
2|\mathcal{S}| \exp\left(-2N_s(t_n) \cdot \frac{1}{16|\mathcal{S}|^2 (x_{\max}+2)^2} \cdot \frac{L \cdot \log(t_n)}{N_s(t_n)}) \right) \vspace{0.2cm}\\ 
\leq 2|\mathcal{S}| t_n^{-1}, \vspace{0.2cm}\\
$
where the last inequality is due to (\ref{eq:L}), and the fact that $\forall i: N_s(t) > T_s^i(t)$.
The first term in (\ref{eq:app15}) is bounded similarly, and therefore:
\begin{align}
\label{eq:app23}
Pr[i,s,n] \leq 2(\frac{|\mathcal{S}||\mathcal{X}_{\max}|}{\pi_{\min}}+2|\mathcal{S}|)\cdot t_n^{-1}.
\end{align} 
Using (\ref{eq:app23}), we can bound (\ref{eq:time_bad_exploit}), and therefore:
\begin{align}
\label{eq:app24} \nonumber
\mathbb{E}[T^i_I(t)] &\leq 6 |\mathcal{S}| \cdot (\frac{|\mathcal{S}||\mathcal{X}_{\max}|}{\pi_{\min}}+2|\mathcal{S}|) \cdot \max_s \pi_s \\ &\cdot \lceil \log_4(\frac{3}{2}t+1) \rceil.
\end{align}

\section*{Appendix E}
\label{App.E}
\section*{Incorporating the regret events to prove Theorem \ref{th:regret}}
We conclude the proof of Theorem \ref{th:regret} by incorporating the regret events discussed above., i.e., regret that is caused by imprecise estimation of the exploration rate, regret that is caused by playing bad arms in  exploration phases, and regret that is caused by playing bad arms in exploitation phases.

Lemmas \ref{lemma:T1} and  \ref{lemma:exp_regret_phi} result in the additional constant term $O(1)$ in the regret bound  (\ref{eq:regret}) which is independent of $t$.

From Lemmas \ref{lemma:exploration} and  \ref{lemma:exp_regret_phi}, the regret caused by playing bad arms in exploration phases by time $t$ is bounded by: 
\begin{flalign*}
 &x_{\max} \cdot \sum \limits_{i=1}^N 
\bigg[\frac{1}{3} [4(3A_i\cdot \log(t)+1)-1]\nonumber\\
&\vspace{0.3cm} \hspace{3cm}  +M^i_{\max} \cdot \log_4(3A_i\log(t)+1) \bigg], 
\end{flalign*}
which coincides with the first and second terms on the RHS of (\ref{eq:regret}).

From Lemma \ref{lemma:exploitation} and Lemma \ref{lemma:exp_regret_phi}, the regret caused by playing sub-optimal arms in exploitation phases by time $t$ is bounded by: 
\[
\:\:x_{\max} \cdot N \cdot 6 |\mathcal{S}| \cdot (\frac{|\mathcal{S}||\mathcal{X}_{\max}|}{\pi_{\min}}+2|\mathcal{S}|) \cdot \max_s \pi_s \cdot \lceil \log_4(\frac{3}{2}t+1) \rceil,\]
which coincides with the third term on the RHS of (\ref{eq:regret}), and thus Theorem \ref{th:regret} follows.

\bibliographystyle{ieeetr}

\end{document}

%% file: macros.tex
\newtheorem{theorem}{Theorem}
\newtheorem{lemma}{Lemma}

\newtheorem{definition}{Definition}

\newcommand{\beq}{\begin{equation}}
\newcommand{\eeq}{\end{equation}}
\newcommand{\bea}{\begin{array}}
\newcommand{\ena}{\end{array}}
\newcommand{\bds}{\begin {itemize}}
\newcommand{\eds}{\end {itemize}}
\newcommand{\bdf}{\begin{definition}}
\newcommand{\blm}{\begin{lemma}}
\newcommand{\edf}{\end{definition}}
\newcommand{\elm}{\end{lemma}}
\newcommand{\bthm}{\begin{theorem}}
\newcommand{\ethm}{\end{theorem}}
\newcommand{\bprp}{\begin{prop}}
\newcommand{\eprp}{\end{prop}}
\newcommand{\bcl}{\begin{claim}}
\newcommand{\ecl}{\end{claim}}
\newcommand{\bcr}{\begin{coro}}
\newcommand{\ecr}{\end{coro}}
\newcommand{\bquest}{\begin{question}}
\newcommand{\equest}{\end{question}}

\newcommand{\larrow}{{\larrow}}

\newcommand{\nin}{{\not \in}}

\def\urltilda{\kern -.15em\lower .7ex\hbox{\~{}}\kern .04em}

%% file: Ga_Ye_Co_Restless.bbl
\begin{thebibliography}{10}

\bibitem{gafni2022Exogenous}
T.~Gafni, M.~Yemini, and K.~Cohen, ``Restless multi-armed bandits under
  exogenous global markov process,'' {\em to appear in IEEE International
  Conference on Acoustics, Speech and Signal Processing (ICASSP), May 2022.}

\bibitem{sun2021dynamic}
J.~Sun, Y.~Zhao, N.~Zhang, X.~Chen, Q.~Hu, and J.~Song, ``A dynamic distributed
  energy storage control strategy for providing primary frequency regulation
  using multi-armed bandits method,'' {\em IET Generation, Transmission \&
  Distribution}, 2021.

\bibitem{scott2015multi}
S.~L. Scott, ``Multi-armed bandit experiments in the online service economy,''
  {\em Applied Stochastic Models in Business and Industry}, vol.~31, no.~1,
  pp.~37--45, 2015.

\bibitem{bulucu2019personalizing}
C.~Bulucu, {\em Personalizing treatments via contextual multi-armed bandits by
  identifying relevance}.
\newblock PhD thesis, Bilkent University, 2019.

\bibitem{hsu2019scheduling}
Y.-P. Hsu, E.~Modiano, and L.~Duan, ``Scheduling algorithms for minimizing age
  of information in wireless broadcast networks with random arrivals,'' {\em
  IEEE Transactions on Mobile Computing}, vol.~19, no.~12, pp.~2903--2915,
  2019.

\bibitem{gafni2022federated}
T.~Gafni, N.~Shlezinger, K.~Cohen, Y.~C. Eldar, and H.~V. Poor, ``Federated
  learning: A signal processing perspective,'' {\em IEEE Signal Processing
  Magazine}, vol.~39, no.~3, pp.~14--41, 2022.

\bibitem{xu2021task}
Y.~Xu, P.~Cheng, Z.~Chen, M.~Ding, Y.~Li, and B.~Vucetic, ``Task offloading for
  large-scale asynchronous mobile edge computing: An index policy approach,''
  {\em IEEE Trans. Signal Process.}, vol.~69, pp.~401--416, 2021.

\bibitem{amar2021online}
O.~Amar and K.~Cohen, ``Online learning for shortest path and backpressure
  routing in wireless networks,'' in {\em IEEE International Symposium on
  Information Theory (ISIT)}, pp.~2702--2707, 2021.

\bibitem{zhao2007survey}
Q.~Zhao and B.~M. Sadler, ``A survey of dynamic spectrum access,'' {\em IEEE
  signal processing magazine}, vol.~24, no.~3, pp.~79--89, 2007.

\bibitem{geirhofer2006measurement}
S.~Geirhofer, L.~Tong, and B.~M. Sadler, ``A measurement-based model for
  dynamic spectrum access in wlan channels,'' in {\em MILCOM 2006-2006 IEEE
  Military Communications conference}, pp.~1--7, IEEE, 2006.

\bibitem{wang2011optimality}
K.~Wang and L.~Chen, ``On optimality of myopic policy for restless multi-armed
  bandit problem: An axiomatic approach,'' {\em IEEE Trans. Signal Process.},
  vol.~60, no.~1, pp.~300--309, 2011.

\bibitem{wang1995finite}
H.~S. Wang and N.~Moayeri, ``Finite-state markov channel-a useful model for
  radio communication channels,'' {\em IEEE Trans. Veh. Technol.}, vol.~44,
  no.~1, pp.~163--171, 1995.

\bibitem{elena2021survey}
G.~Elena, K.~Milos, and I.~Eugene, ``Survey of multiarmed bandit algorithms
  applied to recommendation systems,'' {\em International Journal of Open
  Information Technologies}, vol.~9, no.~4, pp.~12--27, 2021.

\bibitem{kulkarni2020context}
S.~Kulkarni and S.~F. Rodd, ``Context aware recommendation systems: A review of
  the state of the art techniques,'' {\em Computer Science Review}, vol.~37,
  p.~100255, 2020.

\bibitem{misra2019dynamic}
K.~Misra, E.~M. Schwartz, and J.~Abernethy, ``Dynamic online pricing with
  incomplete information using multiarmed bandit experiments,'' {\em Marketing
  Science}, vol.~38, no.~2, pp.~226--252, 2019.

\bibitem{papadimitriou1994complexity}
C.~H. Papadimitriou and J.~N. Tsitsiklis, ``The complexity of optimal queueing
  network control,'' in {\em Proceedings of IEEE 9th Annual Conference on
  Structure in Complexity Theory}, pp.~318--322, 1994.

\bibitem{anantharam1987asymptotically}
V.~Anantharam, P.~Varaiya, and J.~Walrand, ``Asymptotically efficient
  allocation rules for the multiarmed bandit problem with multiple plays-part
  ii: Markovian rewards,'' {\em IEEE Trans. Autom. Control}, vol.~32, no.~11,
  pp.~977--982, 1987.

\bibitem{tekin2012online}
C.~Tekin and M.~Liu, ``Online learning of rested and restless bandits,'' {\em
  IEEE Trans. Inf. Theory}, vol.~58, no.~8, pp.~5588--5611, 2012.

\bibitem{liu2013learning}
H.~Liu, K.~Liu, and Q.~Zhao, ``Learning in a changing world: Restless
  multiarmed bandit with unknown dynamics,'' {\em IEEE Trans. Inf. Theory},
  vol.~3, no.~59, pp.~1902--1916, 2013.

\bibitem{gafni2020learning}
T.~Gafni and K.~Cohen, ``Learning in restless multi-armed bandits via adaptive
  arm sequencing rules,'' {\em IEEE Trans. Autom. Control}, vol.~66, no.~10,
  pp.~5029--5036, 2020.

\bibitem{dai2011non}
W.~Dai, Y.~Gai, B.~Krishnamachari, and Q.~Zhao, ``The non-bayesian restless
  multi-armed bandit: A case of near-logarithmic regret,'' in {\em 2011 IEEE
  International Conference on Acoustics, Speech and Signal Processing
  (ICASSP)}, pp.~2940--2943, 2011.

\bibitem{bagheri2015restless}
S.~Bagheri and A.~Scaglione, ``The restless multi-armed bandit formulation of
  the cognitive compressive sensing problem,'' {\em IEEE Trans. Signal
  Process.}, vol.~63, no.~5, pp.~1183--1198, 2015.

\bibitem{auer2002nonstochastic}
P.~Auer, N.~Cesa-Bianchi, Y.~Freund, and R.~E. Schapire, ``The nonstochastic
  multiarmed bandit problem,'' {\em SIAM journal on computing}, vol.~32, no.~1,
  pp.~48--77, 2002.

\bibitem{garivier2011upper}
A.~Garivier and E.~Moulines, ``On upper-confidence bound policies for switching
  bandit problems,'' in {\em International Conference on Algorithmic Learning
  Theory}, pp.~174--188, Springer, 2011.

\bibitem{audibert2010best}
J.-Y. Audibert, S.~Bubeck, and R.~Munos, ``Best arm identification in
  multi-armed bandits.,'' in {\em COLT}, pp.~41--53, Citeseer, 2010.

\bibitem{shahrampour2017sequential}
S.~Shahrampour, M.~Noshad, and V.~Tarokh, ``On sequential elimination
  algorithms for best-arm identification in multi-armed bandits,'' {\em IEEE
  Trans. Signal Process.}, vol.~65, no.~16, pp.~4281--4292, 2017.

\bibitem{shen2019universal}
C.~Shen, ``Universal best arm identification,'' {\em IEEE Trans. Signal
  Process.}, vol.~67, no.~17, pp.~4464--4478, 2019.

\bibitem{gittins1979bandit}
J.~C. Gittins, ``Bandit processes and dynamic allocation indices,'' {\em
  Journal of the Royal Statistical Society: Series B (Methodological)},
  vol.~41, no.~2, pp.~148--164, 1979.

\bibitem{lai1985asymptotically}
T.~L. Lai and H.~Robbins, ``Asymptotically efficient adaptive allocation
  rules,'' {\em Advances in applied mathematics}, vol.~6, no.~1, pp.~4--22,
  1985.

\bibitem{auer2002finite}
P.~Auer, N.~Cesa-Bianchi, and P.~Fischer, ``Finite-time analysis of the
  multiarmed bandit problem,'' {\em Machine learning}, vol.~47, no.~2,
  pp.~235--256, 2002.

\bibitem{tekin2010online}
C.~Tekin and M.~Liu, ``Online algorithms for the multi-armed bandit problem
  with markovian rewards,'' in {\em 2010 48th Annual Allerton Conference on
  Communication, Control, and Computing (Allerton)}, pp.~1675--1682, 2010.

\bibitem{tekin2012approximately}
C.~Tekin and M.~Liu, ``Approximately optimal adaptive learning in opportunistic
  spectrum access,'' in {\em 2012 Proceedings IEEE INFOCOM}, pp.~1548--1556,
  2012.

\bibitem{xu2021online}
J.~Xu, L.~Chen, and O.~Tang, ``An online algorithm for the risk-aware restless
  bandit,'' {\em European Journal of Operational Research}, vol.~290, no.~2,
  pp.~622--639, 2021.

\bibitem{karthik2021learning}
P.~Karthik and R.~Sundaresan, ``Learning to detect an odd restless markov
  arm,'' in {\em 2021 IEEE International Symposium on Information Theory
  (ISIT)}, pp.~1457--1462, 2021.

\bibitem{gafni2019distributed}
T.~Gafni and K.~Cohen, ``A distributed stable strategy learning algorithm for
  multi-user dynamic spectrum access,'' in {\em 2019 57th Annual Allerton
  Conference on Communication, Control, and Computing (Allerton)},
  pp.~347--351, 2019.

\bibitem{gafni2021distributed}
T.~Gafni and K.~Cohen, ``Distributed learning over markovian fading channels
  for stable spectrum access,'' {\em arXiv preprint arXiv:2101.11292}, 2021.

\bibitem{whittle1988restless}
P.~Whittle, ``Restless bandits: Activity allocation in a changing world,'' {\em
  J. Appl. Probab.}, vol.~25, no.~A, pp.~287--298, 1988.

\bibitem{weber1990index}
R.~R. Weber and G.~Weiss, ``On an index policy for restless bandits,'' {\em J.
  Appl. Probab.}, vol.~27, no.~3, pp.~637--648, 1990.

\bibitem{ehsan2004optimality}
N.~Ehsan and M.~Liu, ``On the optimality of an index policy for bandwidth
  allocation with delayed state observation and differentiated services,'' in
  {\em IEEE INFOCOM 2004}, vol.~3, pp.~1974--1983, 2004.

\bibitem{cohen2014restless}
K.~Cohen, Q.~Zhao, and A.~Scaglione, ``Restless multi-armed bandits under
  time-varying activation constraints for dynamic spectrum access,'' in {\em
  2014 48th Asilomar Conference on Signals, Systems and Computers},
  pp.~1575--1578, 2014.

\bibitem{zhao2008myopic}
Q.~Zhao, B.~Krishnamachari, and K.~Liu, ``On myopic sensing for multi-channel
  opportunistic access: structure, optimality, and performance,'' {\em IEEE
  Trans. Wireless Commun.}, vol.~7, no.~12, pp.~5431--5440, 2008.

\bibitem{ahmad2009optimality}
S.~H.~A. Ahmad, M.~Liu, T.~Javidi, Q.~Zhao, and B.~Krishnamachari, ``Optimality
  of myopic sensing in multichannel opportunistic access,'' {\em IEEE Trans.
  Inf. Theory}, vol.~55, no.~9, pp.~4040--4050, 2009.

\bibitem{ahmad2009multi}
S.~H.~A. Ahmad and M.~Liu, ``Multi-channel opportunistic access: A case of
  restless bandits with multiple plays,'' in {\em 2009 47th Annual Allerton
  Conference on Communication, Control, and Computing (Allerton)},
  pp.~1361--1368, 2009.

\bibitem{liu2010indexability}
K.~Liu and Q.~Zhao, ``Indexability of restless bandit problems and optimality
  of whittle index for dynamic multichannel access,'' {\em IEEE Trans. Inf.
  Theory}, vol.~56, no.~11, pp.~5547--5567, 2010.

\bibitem{wang2013optimality}
K.~Wang, L.~Chen, and Q.~Liu, ``On optimality of myopic policy for
  opportunistic access with nonidentical channels and imperfect sensing,'' {\em
  IEEE Trans. Veh. Technol.}, vol.~63, no.~5, pp.~2478--2483, 2013.

\bibitem{zhao2007structure}
Q.~Zhao and B.~Krishnamachari, ``Structure and optimality of myopic sensing for
  opportunistic spectrum access,'' in {\em 2007 IEEE International Conference
  on Communications}, pp.~6476--6481, 2007.

\bibitem{liu2011indexability}
K.~Liu, R.~Weber, and Q.~Zhao, ``Indexability and whittle index for restless
  bandit problems involving reset processes,'' in {\em 2011 50th IEEE
  Conference on Decision and Control and European Control Conference},
  pp.~7690--7696, 2011.

\bibitem{krishnamurthy2009partially}
V.~Krishnamurthy and B.~Wahlberg, ``Partially observed markov decision process
  multiarmed bandits—structural results,'' {\em Mathematics of Operations
  Research}, vol.~34, no.~2, pp.~287--302, 2009.

\bibitem{krishnamurthy2001hidden}
V.~Krishnamurthy and R.~J. Evans, ``Hidden markov model multiarm bandits: a
  methodology for beam scheduling in multitarget tracking,'' {\em IEEE Trans.
  Signal Process.}, vol.~49, no.~12, pp.~2893--2908, 2001.

\bibitem{hartland2006multi}
C.~Hartland, S.~Gelly, N.~Baskiotis, O.~Teytaud, and M.~Sebag, ``Multi-armed
  bandit, dynamic environments and meta-bandits", in \textit{nIPS-2006
  Workshop, Online Trading Between Exploration and Exploitation, Whistler,
  Canada},'' 2006.

\bibitem{yu2009piecewise}
J.~Y. Yu and S.~Mannor, ``Piecewise-stationary bandit problems with side
  observations,'' in {\em Proceedings of the 26th annual international
  conference on machine learning}, pp.~1177--1184, 2009.

\bibitem{slivkins2008adapting}
A.~Slivkins and E.~Upfal, ``Adapting to a changing environment: the brownian
  restless bandits.,'' in {\em COLT}, pp.~343--354, 2008.

\bibitem{wang2018regional}
Z.~Wang, R.~Zhou, and C.~Shen, ``Regional multi-armed bandits with partial
  informativeness,'' {\em IEEE Trans. Signal Process.}, vol.~66, no.~21,
  pp.~5705--5717, 2018.

\bibitem{baltaoglu2016online}
S.~Baltaoglu, L.~Tong, and Q.~Zhao, ``Online learning and optimization of
  markov jump linear models,'' in {\em 2016 IEEE International Conference on
  Acoustics, Speech and Signal Processing (ICASSP)}, pp.~2289--2293, 2016.

\bibitem{baltaoglu2016onlinea}
S.~Baltaoglu, L.~Tong, and Q.~Zhao, ``Online learning and optimization of
  markov jump affine models,'' {\em arXiv preprint arXiv:1605.02213}, 2016.

\bibitem{yemini2021restless}
M.~Yemini, A.~Leshem, and A.~Somekh-Baruch, ``The restless hidden markov bandit
  with linear rewards and side information,'' {\em IEEE Trans. Signal
  Process.}, vol.~69, pp.~1108--1123, 2021.

\bibitem{al2018accurate}
A.~Al-Tahmeesschi, M.~L{\'o}pez-Ben{\'\i}tez, J.~Lehtom{\"a}ki, and
  K.~Umebayashi, ``Accurate estimation of primary user traffic based on
  periodic spectrum sensing,'' in {\em 2018 IEEE Wireless Communications and
  Networking Conference (WCNC)}, pp.~1--6, IEEE, 2018.

\bibitem{auer2008near}
P.~Auer, T.~Jaksch, and R.~Ortner, ``Near-optimal regret bounds for
  reinforcement learning,'' {\em Advances in neural information processing
  systems}, vol.~21, 2008.

\bibitem{lezaud1998chernoff}
P.~Lezaud, ``Chernoff-type bound for finite markov chains,'' {\em Annals of
  Applied Probability}, pp.~849--867, 1998.

\bibitem{liu2008link}
K.~Liu and Q.~Zhao, ``Link throughput of multi-channel opportunistic access
  with limited sensing,'' in {\em 2008 IEEE International Conference on
  Acoustics, Speech and Signal Processing}, pp.~2997--3000, IEEE, 2008.

\end{thebibliography}
